\def\cvprPaperID{0000}
\def\confName{CVPR}
\def\confYear{2023}

\def\authorBlock{
    Jing Wang$^1$ \thanks{Equal contribution} \qquad
    Songtao Wu$^1$ \footnotemark[1] \qquad
    Kuanhong Xu$^1$ \qquad
    Zhiqiang Yuan$^2$
     \\
    $^1$Sony R\& D Center China, Beijing Lab \\
    $^2$Aerospace Information Research Institute, Chinese Academy of Science\\
    {\tt\small \{JingD.Wang, Songtao.Wu, Kuanhong.Xu\}@sony.com, yuanzhiqiang19@mails.ucas.ac.cn}
}

\newif\ifreview 
\newif\ifarxiv 
\newif\ifcamera 
\newif\ifrebuttal 

\pdfoutput=1
\documentclass[10pt,twocolumn,letterpaper]{article}
\ifreview \usepackage[review]{cvpr} \fi
\ifarxiv \usepackage[pagenumbers]{cvpr} \fi
\ifrebuttal \usepackage[rebuttal]{cvpr} \fi
\ifcamera \usepackage{cvpr} \fi

\usepackage{graphicx}
\usepackage{amsmath}
\usepackage{amssymb}
\usepackage{booktabs}

\usepackage{times}
\usepackage{epsfig}
\usepackage{graphicx}
\usepackage{bbm}
\usepackage{amsfonts}
\usepackage{mathrsfs}
\usepackage{multirow}
\usepackage{ulem}

\usepackage{algorithm}
\usepackage{algpseudocode}

\usepackage{times}
\usepackage{microtype}
\usepackage{epsfig}
\usepackage[table,xcdraw]{xcolor}
\usepackage{subcaption}
\usepackage{caption}
\usepackage{float}
\usepackage{placeins}
\usepackage{color, colortbl}
\usepackage{stfloats}
\usepackage{enumitem}
\usepackage{tabularx}
\usepackage{xstring}
\usepackage{multirow}
\usepackage{xspace}
\usepackage{url}
\usepackage{xcolor}
\usepackage[hang,flushmargin]{footmisc}

\ifcamera \usepackage[accsupp]{axessibility} \fi

% ...

\ifarxiv  \fi

\newcommand{\R}[1]{{%
    \textbf{%
        \ifstrequal{#1}{1}{\textcolor{red}{R#1}}{%
        \ifstrequal{#1}{2}{\textcolor{blue}{R#1}}{%
        \ifstrequal{#1}{3}{\textcolor{magenta}{R#1}}{%
        \ifstrequal{#1}{4}{\textcolor{teal}{R#1}}{%
                           \textcolor{cyan}{R#1}%
        }}}}%
    }%
}}

\usepackage{xr-hyper}

\makeatletter
\newcommand*{\addFileDependency}[1]{
  \typeout{(#1)}
  \@addtofilelist{#1}
  \IfFileExists{#1}{}{\typeout{No file #1.}}
}

\makeatother

\usepackage[pagebackref,breaklinks,colorlinks]{hyperref}
\usepackage[capitalize]{cleveref}
\crefname{section}{Sec.}{Secs.}
\crefname{table}{Table}{Tables}
\crefname{figure}{Fig.}{Figs.}

\begin{document}
\title{Frequency Compensated Diffusion Model for Real-scene Dehazing}
\author{\authorBlock} %
\maketitle

\begin{abstract}
Due to distribution shift, deep learning based methods for image dehazing suffer from performance degradation when applied to real-world hazy images. In this paper, we consider a dehazing framework based on conditional diffusion models for improved generalization to real haze. First, we find that optimizing the training objective of diffusion models, i.e., Gaussian noise vectors, is non-trivial. The spectral bias of deep networks hinders the higher frequency modes in Gaussian vectors from being learned and hence impairs the reconstruction of image details. To tackle this issue, we design a network unit, named Frequency Compensation block (FCB), with a bank of filters that jointly emphasize the mid-to-high frequencies of an input signal. We demonstrate that diffusion models with FCB achieve significant gains in both perceptual and distortion metrics. Second, to further boost the generalization performance, we propose a novel data synthesis pipeline, \textit{HazeAug}, to augment haze in terms of \textit{degree} and \textit{diversity}. Within the framework, a solid baseline for blind dehazing is set up where models are trained on synthetic hazy-clean pairs, and directly generalize to real data. Extensive evaluations show that the proposed dehazing diffusion model significantly outperforms state-of-the-art methods on real-world images. Our code is at \href{https://github.com/W-Jilly/frequency-compensated-diffusion-model-pytorch}{https://github.com/W-Jilly/frequency-compensated-diffusion-model-pytorch}.
\end{abstract}
\section{Introduction}
\label{sec:intro}

Single image dehazing aims to recover the clean image from a hazy input for improved visual quality. The hazing process is formulated by the well-known atmospheric scattering model (ASM) \cite{nayar1999vision}: 
\begin{equation}
  I(p) = J(p) e^{-\beta d(p)} + A(1-e^{-\beta d(p)})
  \label{eq:hazy-equation}
\end{equation}
\noindent
where $p$ is the pixel position, $I(p)$ and $J(p)$ are the hazy and haze-free image respectively. $A$ denotes the global atmospheric light. $e^{-\beta d(p)}$ is the transmission map, in which $d(p)$ is the scene depth and the scattering coefficient $\beta$ represents the degree of haze.

Existing dehazing algorithms have been impressively advanced by the emergence of deep neural networks (DNNs) \cite{fattal2014dehazing, gcanet, FFA, MSBDN}, but generalize poorly to real-world hazy images. An important reason is that training in a supervised manner requires hazy-clean pairs in large amounts. Due to the inaccessibility of real paired data, synthetic pairs created from ASM are widely adopted for haze simulation. Unfortunately, current ASM-based synthesis lacks ability to faithfully represent real hazy conditions. For example, $\beta$ and $A$, modeled as global constants in Eq.\ref{eq:hazy-equation}, are actually found to be pixel position dependent under some conditions \cite{wang2021fwb, wang2021fully}, \textit{i.e.,} non-homogeneous ASM. Consequently, dehazing models trained on synthetic pairs work well in-distribution, but degrades dramatically when applied to real haze.
To better address the problem of dehazing in real scene, many works \cite{shao2020domain, PSD} are carried out within a paradigm of semi-supervised learning that assumes the availability of some real images. Domain adaptation techniques are often utilized to bridge the gap between synthetic and real data. However, the generality of DNNs is not effectively improved, as one needs to re-train the model once encountering a new scenario.

Recently, diffusion models \cite{DDPM-song, DDPM-theory, ho2020denoising}, belonging to a family of deep generative models, emerged with impressive results that beat Generative Adversarial Network (GAN) \cite{goodfellow2020generative} baselines on image generation.
Inspired by their strong ability in modeling data distributions, there are attempts to learn the prior distribution of groundtruth images alone with diffusion models for inverse problem solving \cite{Sdedit, song2021solving}. In other words, the data prior is learned by unconditional diffusion models, whereas only in the generative process, an input (\textit{e.g.,} hazy image) "hijacks" to guide the image generation to be consistent with the input and the prior. Despite being an universal approach, the missing of task-specific knowledge at training usually leads to an inferior performance to conditional models. To tackle this issue, conditional diffusion models are proposed to make the training of networks accept a conditioning input \cite{sr3} and demonstrate remarkable performance on a variety of image-to-image applications, \textit{e.g.,} super-resolution \cite{sr3} and deblurring \cite{Whang_2022_CVPR}.

\begin{figure*}[!t]
\centering
\includegraphics [width=6.8in]{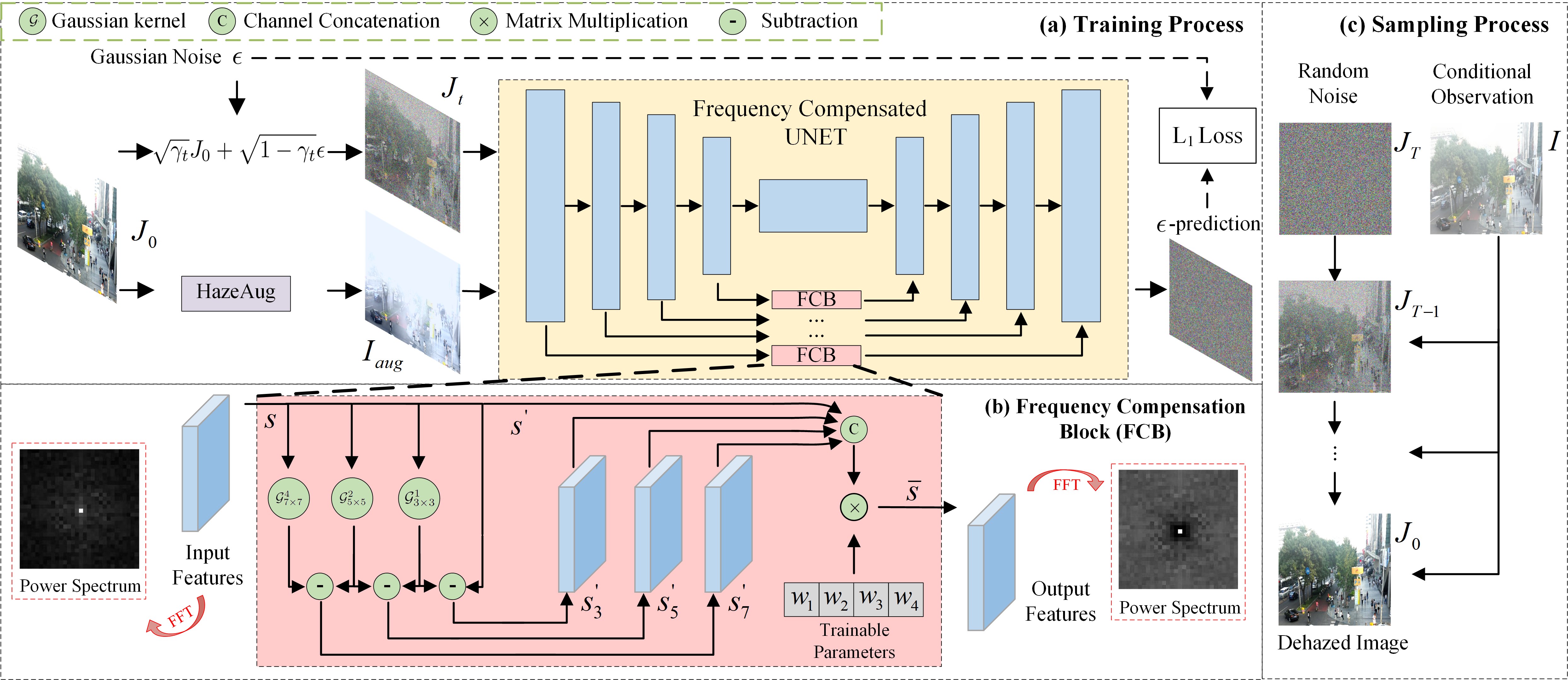}
\caption{(a) The training process of the proposed dehazing diffusion model. At step $t$, the network takes an augmented hazy image $I_{aug}$ and a noisy image $J_t$ as inputs. The network architecture adopts special skip connections, \textit{i.e.,} the Frequency Compensation Block (FCB), for better $\epsilon$-prediction. (b) The detailed block design of FCB. The input signals of FCB are enhanced at the mid-to-high frequency band so that the output spectrum has abundant higher frequency modes. (c) The sampling process of the proposed dehazing diffusion model.
}
\label{framework}
\end{figure*}

Motivated by their promising results, 
we seek to adapt the conditional diffusion models 
for dehazing task. It has been reported that diffusion models, on various applications \cite{sr3, Whang_2022_CVPR, Palette}, deliver high perceptual quality outputs, but often underperform in terms of distortion measures, \textit{e.g.,} PSNR and SSIM, which suggests the reconstructed image details are not very well aligned. In this work, we find this to be closely related to a property of neural networks, namely spectral bias that prioritizes learning the low-frequency functions in the training \cite{rahaman2019spectral}. However, this property is not aligned with the learning objective of diffusion models, which resembles estimating Gaussian noise vectors with uniformly distributed spectra. Their abundant high frequency modes are hence harder and slower to be learned due to the bias, which accordingly, impairs the restored high-frequency image details. To fill this gap, we propose a generic network unit, named Frequency Compensation block (FCB), to allow selective emphasis on the mid-to-high frequencies of an input signal. The block structure (see Fig.\ref{framework}) mimics the filter design in signal processing through which only the frequency band of interest is allowed to pass. It can be plugged into any encoder-decoder architecture to ease the learning of an unnatural Gaussian-like target. Experiments validate the effectiveness of FCB in improving both the perceptual quality and distortion measures of diffusion models.

Next, we propose a novel data augmentation pipeline, HazeAug, to improve the generalization ability of diffusion models for "blind" dehazing. In a blind setting, the dehazing model is trained on synthetic pairs only, and directly applied to real hazy images. We hypothesize that the diffusion models possess much stronger generalization ability if trained on \textit{richer} data. However, ASM-based synthesis is limited in its ability to describe diverse and complex hazy conditions. To address this limitation, HazeAug aims to augment haze in terms of \textit{degree} and \textit{diversity}. Specifically, HazeAug generates more hard samples of severe hazy conditions and amplifies haze diversity by migrating haze patterns from one image to another. Train-set thus get effectively enriched beyond ASM to capture a broader slice of real data distribution. As a result, we demonstrate that diffusion models trained on synthetic haze only can outperform many state-of-the-art real scene dehazing methods that require prior knowledge of real haze. Interestingly, diffusion models are empirically found to benefit more from augmentation than other deep learning based methods, owing largely to their strong ability in data modeling. 


In summary, our main contributions include:
\begin{itemize}
    \item We set up a solid baseline with a conditional diffusion model based framework for ``blind'' dehazing. Blind dehazing is to train with synthetic data only, and directly transfer to real haze. Our framework demonstrates superior generalization performance on real data.
    \item We design the Frequency Compensation block, a novel architectural unit to ease the optimization of Gaussian-like target of diffusion models. It significantly improves the performance of diffusion models on real hazy images in terms of both perceptual and distortion metrics.
    

    \item A novel data augmentation pipeline, \textit{HazeAug}, is proposed to go beyond ASM and generalize better to unknown scenarios. It enables a dehazing diffusion model to flexibly adapt to real world images with state-of-the-art performance.
\end{itemize}
\section{Related Work}
\label{sec:related}

\subsection{Single Image Dehazing}
Existing dehazing algorithms can be roughly divided into prior-based and learning-based methods.

\textbf{Prior-based methods} rely on hand-crafted priors or assumptions (\textit{e.g.,} scene albedo or transmission map in the scattering model) to recover haze-free image \cite{13, nishino2012bayesian}.
He $et \ al.$ \cite{he2010single} use the dark channel prior (DCP) to estimate the transmission map, assuming that the lowest pixel value should be close to zero in at least one channel except for the sky region.
It inspires a series of work to estimate the transmission map and atmospheric light by a variety of improved priors \cite{fattal2014dehazing, zhu2015fast, bui2017single, xu2012fast, tang2014investigating}.
For example, Fattal \cite{fattal2014dehazing} derive a local formation model that explains the color-lines in hazy scene and leverage it to recover the scene transmission. 
However, these priors are not always reliably estimated in real scenarios, and therefore cause unsatisfactory results.

\textbf{Learning-based methods} refer to learn the intrinsic dehazing process in a data-driven way empowered by deep neural networks \cite{gcanet, did, li2017aod}.
Early studies often use DNNs to predict the intermediate transmission map, and remove haze based on ASM \cite{14, zhang2018densely, zhao2017deep}.
Alternatively, many works adopt a fully end-to-end approach where DNNs are trained to approximate the groundtruth haze-free images \cite{FFA, MSBDN, li2017aod}.
In recent years, a large number of learning-based methods focus on designing better architectures \cite{gcanet, FFA, MSBDN} or improving optimization \cite{liu2019learning, li2019pdr} for dehazing task so as to achieve better performance. 

Approaches based on deep generative models have also been applied to image dehazing. Conditional GANs are usually utilized with a discriminator to constrain the naturalness of the dehazed images \cite{dong2020fd}. In addition, CycleGAN \cite{zhu2017unpaired} tackles unpaired image setting and can be used to learn to dehaze without paired data \cite{engin2018cycle, liu2020end, li2017dr}.
However, despite the remarkable performance achieved by learning-based methods, their data-driven nature makes them highly domain-specific and hard to generalize to real data.

\subsection{Diffusion Models}
Diffusion and score-based models \cite{DDPM-song, DDPM-theory, ho2020denoising}, belonging to a family of deep generative models, result in key advances in many applications such as image generation \cite{dhariwal2021diffusion} and image synthesis \cite{Sdedit}.
Specifically, for controllable image generation, Dhariwal and Nichol \cite{dhariwal2021diffusion} inject class information by training a separate classifier for conditional image synthesis. Saharia $et \ al.$ \cite{saharia2021image} propose the conditional diffusion models to make the denoising process of diffusion models conditioned on an input signal. Moreover, Ho and Salimans \cite{ho2022classifier} jointly train the unconditional and conditional models to avoid any extra classifiers. So far, we have seen their success on a broad set of tasks, \textit{e.g.,} deblurring \cite{Whang_2022_CVPR}, inpainting \cite{Palette} and image editing \cite{avrahami2022blended}. Our work is based on \cite{saharia2021image} and to the best of our knowledge, we are the first to apply conditional diffusion models to dehazing task and achieve the superior performance on real hazy images.

\section{Methodology}
\label{sec:method}

\subsection{Conditional Denoising Diffusion Model}
Diffusion models, such as denoising diffusion probabilistic models (DDPMs) \cite{ho2020denoising}, progressively corrupt data by injecting noise and then learn its reverse process for sample generation. The sampling procedure of DDPMs starts from a Gaussian noise, and produces an image via a Markov chain of iterative denoising. For conditional image generation, the learning of reverse process can be further adapted to take an extra conditional input \cite{saharia2021image}. In this section we briefly review the training and sampling of conditional DDPMs in the context of image dehazing. One can refer to \cite{DDPM-theory, ho2020denoising} for detailed derivation of diffusion models.

Given a clean image $J_0$ (or $J$), it is repeatedly corrupted by adding Gaussian noise over $T$ steps in the forward process.
We can obtain its noisy version $J_t$ at step $t$ by sampling a vector $\epsilon \sim \mathcal{N}(\mathbf{0}, \mathbf{I})$ and applying the following transformation:
\begin{equation}
\begin{split}
    J_t  = 
    \sqrt{\gamma_t}J_0 + \sqrt{1-\gamma_t}\epsilon
\end{split}
\label{eqa:diffusion-noise}
\end{equation}
\noindent
where $\gamma_t=\prod_{i=1}^t \alpha_i$, and $\{\alpha_t \in (0,1) \}_{t=1}^T$ is a sequence controlling the noise scale at every step $t$.

If parameterizing the reverse Markov chain according to \cite{ho2020denoising}, training the reverse process is the same as training a neural network 
$f_\theta$ to predict $\epsilon$ given $J_t$ and $\gamma_t$. For conditional dehazing, $f_\theta$ takes an extra hazy image $I$ as input and the loss based on $L_{1}$ norm is given by:
\begin{equation}
    \mathbb{E}_{(I,J)}\mathbb{E}_{\epsilon \sim \mathcal{N}(0,1),\gamma_t} \lVert f_\theta( I,  \underbrace{\sqrt{\gamma_t} J_0+ \sqrt{1-\gamma_t} \epsilon}_{J_t}, \gamma_t ) - \epsilon \rVert_1^1
\label{eqa:diffusion-loss}
\end{equation}
During sampling, one can gradually dehaze an image $J_{t-1}$ from $J_t$ for $t=T,...,1$ via the updating rule:
\begin{equation}
    J_{t-1} =\frac{1}{\sqrt{\alpha_t}}(J_t - \frac{1-\alpha_t}{\sqrt{1-\gamma_t}}f_\theta(I, J_t, \gamma_t)) 
     + \sqrt{1-\alpha_t} \bar{\epsilon_t}
\end{equation}
\noindent
where $\bar{\epsilon_t} \sim \mathcal{N}(\mathbf{0}, \mathbf{I})$ is the noise in sampling process.

\subsection{Frequency Compensation Block}
\label{section3.2}

Traditional end-to-end methods optimize a data-fit term where a dehazed image is learned to approximate the groundtruth in pixel distance. On the other hand, 
diffusion models are trained via a function approximator $f_\theta$ to predict $\epsilon\sim\mathcal{N}(\mathbf{0}, \mathbf{I})$ in Eq.\ref{eqa:diffusion-loss}. Such fundamental difference in training objectives, if seen in frequency domain, is even more enlightening. The power spectra of natural images (\textit{e.g.,} $J$) tend to fall off with increasing spatial frequency \cite{van1996modelling}, while Gaussian noise is well known to follow a uniform power spectrum density (PSD) \cite{rainal1961sampling, boreman2001modulation}. However, as indicated by \cite{rahaman2019spectral, wang2020high}, DNNs exhibit a bias towards smooth functions that favor low-frequency signals. It is their properties to prioritize fitting low-frequency modes, leaving the high-frequency ones learned late in the training. Such properties align well with the spectra of natural images, but may not be optimal for noise vectors to be learned.

\begin{figure}[!t]
	\centering
	\includegraphics [width=3.0in]{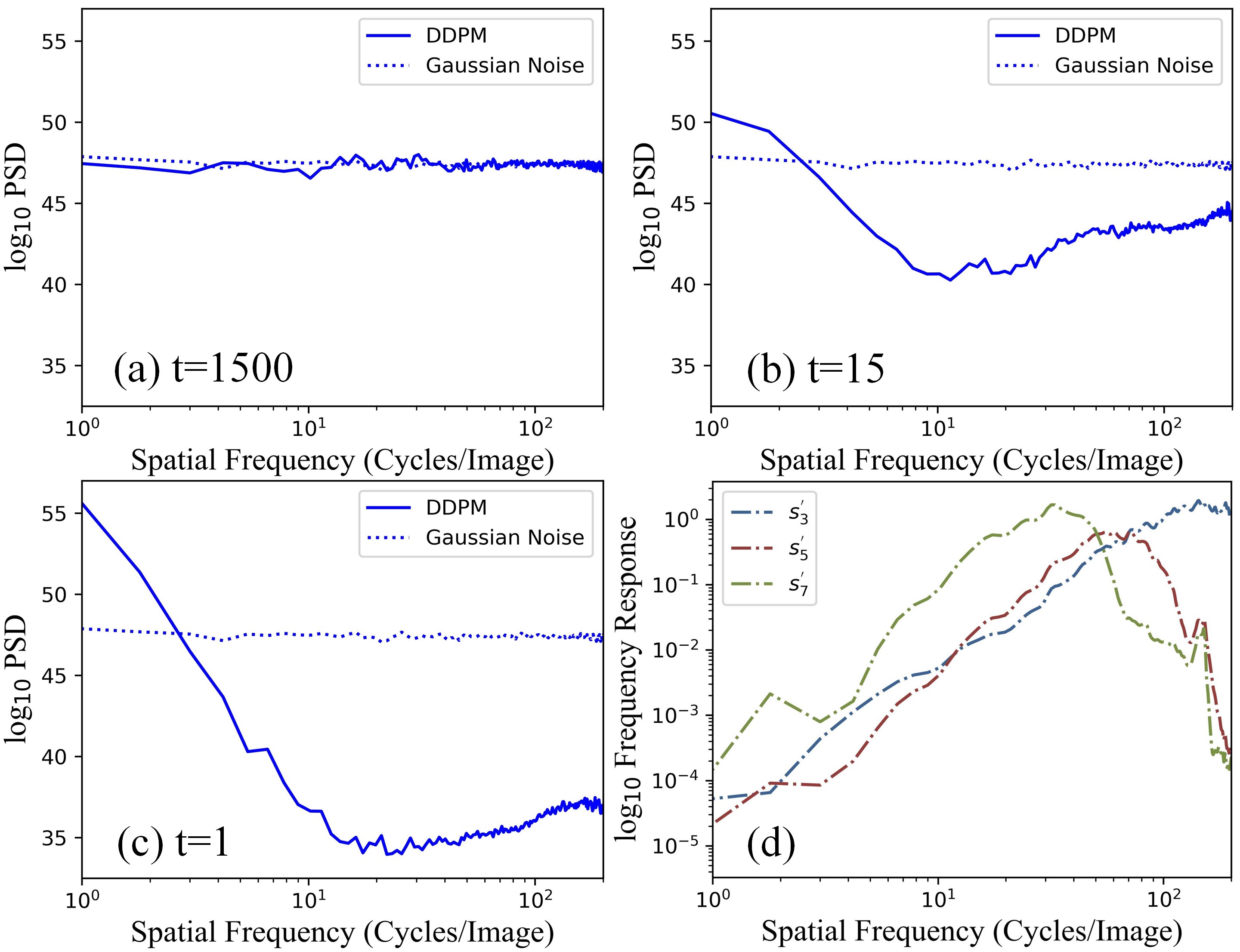}
	\caption{(a)-(c) PSD analysis on the predicted $\epsilon$ of a plain DDPM (solid line) \textit{versus} the groundtruth (dot line). Noise scales $t\in\{1500,15,1\}$ are tested and the results are averaged over train images. (d) Frequency response curves of the band- and high- pass filters within FCB with an example of $M=3$, $k \in \{3, 5, 7\}$ and $\sigma \in \{1, 2, 4\}$. Blue, red and green line is plotted by measuring $s'_{3},s'_{5},s'_{7}$ respectively with the input $s$ stepping through the entire spatial frequency range.}
	\label{power_motivation}
\end{figure}

The aforementioned theoretical findings are also observed in our empirical study. In Fig.\ref{power_motivation}a-c, we firstly train a plain dehazing diffusion model with a U-Net architecture \cite{unet} and perform PSD analysis on the outputs of its final layer with noise corrupted inputs at varying $t$. Though ideally following a uniform PSD, it is actually found in Fig.\ref{power_motivation} that the mid-to-high frequency band gets heavily attenuated as $t$ (\textit{i.e.,} noise scale) becomes small. It is because when noises dominate ($t\rightarrow T$), the input $J_t$ is very close to the target $\epsilon$ (see Eq.\ref{eqa:diffusion-noise}) and the skip connections of U-Net facilitate training by mapping the input identically to the output, hence preserving high frequencies. However, as $t\rightarrow 1$, 
the $\epsilon$-prediction task resembles denoising an image with small amounts of noise while retaining image details.
It is when the insufficiency of networks to capture high-frequency modes becomes profound. This phenomenon at small $t$ in the training strongly relates to the late stage of the sampling process when image details are to be reconstructed. If not well handled, it may potentially result in higher distortion in images and poorer performance.

Therefore, we propose to design a network unit, termed as Frequency Compensation block (FCB), that emphasizes the mid-to-high frequencies in particular. Similar to filter design in signal processing, a FCB consists of a bank of convolutional filters that allows only the mid-to-high frequencies from shallower layers to pass and transfers them to deeper layers via the skip connection structure (see Fig.\ref{framework}). The choice of filters follows two principles:
(1) The filter should be 
radially symmetric to avoid biasing towards the high frequency image features of edges and corners.
(2) The bandwidth of these filters should cover the frequency range where a network finds hard to capture.

For the sake of simplicity, we choose a bank of low-pass Gaussian kernels to compose FCB. The overview of FCB design is in Fig.\ref{framework}b. Let $\mathcal{G}_{k \times k}^{\sigma}$ be a 2D Gaussian kernel of size $k$ and standard deviation $\sigma$. Consider an input feature map of FCB $s \in \mathbb{R}^{H_s \times W_s \times C_s}$ passing through the kernels:
\begin{equation}
   s_{k, \sigma} = \mathcal{G}_{k \times k}^{\sigma} * s, \ \ k \in \{3, 5 , 7 , 9, ...\}
   \label{eqa:fcb-kernel}
\end{equation}
where $s_{k, \sigma}$ is the output of $s$ convolved by $\mathcal{G}_{k \times k}^{\sigma}$. One can choose $M$ filters of different choices of $k$ and $\sigma$ to suit training at varying $t$. Once chosen, they are non-learnable during training to keep the determined bandwidth unchanged. 

Without loss of generality, we take $M=3$ as example and choose the kernels, \textit{i.e.,} $k \in \{3, 5, 7\}$ with some correspondingly user defined $\sigma$. For simplicity, we use $s_{k}$ to denote $s_{k, \sigma}$ in Eq.\ref{eqa:fcb-kernel} for the rest of the paper. 
After calculating the difference between the outputs of two Gaussian kernels, we end up with either band- or high-pass filters \cite{bundy1984difference}:
\begin{equation}
  \mathbf{s}' = [s', s'_{3},s'_{5},s'_{7} ] = [s, s-s_{3},s_{3}-s_{5},s_{5}-s_{7}]
\end{equation}
where $s'_{k}$ denotes the output after passing through a band- or high-pass filter whose frequency response is plotted in Fig.\ref{power_motivation}d. We can see that by carefully selecting Gaussian kernels, one can obtain filters of different frequency bands. These filters can meet the requirements of different $t$ (see Fig.\ref{power_motivation}a-c). When combined, they emphasize the higher end of the frequency range.
Finally, $\mathbf{s}'$ is aggregated via weighted averaging:
\begin{equation}
\bar{\mathbf{s}} = <\mathbf{W},\mathbf{s}'> = w_1 s'+w_2 s'_{3}+w_3 s'_{5}+w_4 s'_{7}
\label{eqa:fcb-agg}
\end{equation}
where $\mathbf{W}=[w_1, w_2, w_3, w_4]$ are trainable weights to be learned to balance the contribution of different filters. Note that in Eq.\ref{eqa:fcb-agg}, we keep an identity mapping of $s$ to $s'$ so as to preserve its original signals. This is especially helpful when training at $t\rightarrow T$ as mentioned above. 

The proposed FCB is to compensate the inadequate mid-to-high frequencies when training DDPMs. In Section \ref{sec:ablation}, we show by experiments that the output spectrum is indeed closer to be uniformly distributed, suggesting the effectiveness of FCB in $\epsilon$-prediction.

\subsection{HazeAug}

Advanced data augmentation is well known to significantly improve the likelihoods of deep generative models \cite{jun2020distribution}. In this section, we show that diffusion models can also benefit from this approach.
To overcome the limitations of current ASM-based synthesis, we propose \textit{HazeAug}, a method that augments haze in terms of \textit{degree} (hard sample) and \textit{diversity} (haze migration) on top of ASM. HazeAug drives a model to learn about invariances to changing hazing process, enabling the trained model to be flexibly adapted to real hazy images without retraining.

Let us consider a synthetic database $\{ (I^{i},J^{i}) \}_{i=1}^{N}$ constructed by current synthesis pipeline \cite{RESIDE} following Eq.\ref{eq:hazy-equation}, where $I^{i},J^{i} \in \mathbb{R}^{H \times W \times 3}$. In Eq.\ref{eq:hazy-equation}, $d(p)$ is obtained from a depth estimator while $A$ and $\beta$ are randomly chosen from some data range. For each clean sample $J^{i}$, HazeAug is applied with either of the following operators to obtain an augmented sample $I_{aug}^{i}$:

\textbf{Hard Sample.}
While $\beta$ and $A$ are usually limited within a small range to ensure only a moderate degree of haze added, we perturb $\beta\in(\beta_{min}, \beta_{max})$ and $A\in(A_{min}, A_{max})$ to a much larger range (details in Appendix \textcolor{red}{A}). Particularly, $\beta_{max}$ and $A_{max}$ are enlarged to include hard samples of very low signal-to-haze ratio, where scene is heavily blocked by strong haze shielding and illumination. The purpose is to cover more challenging real-world cases, \textit{e.g.,} severe hazy conditions. In such cases, the supervisory signals in the conditional input of diffusion models are diminished, whereas the generative modeling ability is emphasized and hence better generality \cite{20}.

\begin{algorithm}[!t]
\caption{\textcolor{black}{The Procedure of HazeAug}}
\label{alg:Framwork}
\begin{algorithmic}[1]
\renewcommand{\algorithmicrequire}{ \textcolor{black}{\textbf{Require:}}}
\Require
\Statex Depth map estimator $d(\cdot)$ 
\Statex Uniform random number generator $U(\cdot,\cdot)$ 
\Statex Thresholds of $A$ and $\beta$: $(A_{min},A_{max})$, $(\beta_{min},\beta_{max})$
\Statex Frequency cutoff threshold: $(\delta_{min},\delta_{max})$
\renewcommand{\algorithmicrequire}{ \textbf{Input:}}
\Require
\Statex Synthetic database $\{ (I^{i},J^{i}) \}_{i=1}^{N}$
\renewcommand{\algorithmicrequire}{ \textbf{For each input sample $J^{i}:$}}
\Require
\State $\varsigma = U(0, 1)$ 
\If{ $\varsigma < 0.5:$ }
\Statex \quad\ \textbf{Hard sample}
\State \quad\ $A = U(A_{min},A_{max})$
\State \quad\ $\beta = U(\beta_{min},\beta_{max})$
\State \quad\ $z^{i} = d(J^{i})$
\State \quad\ $I_{aug}^{i} = J^{i} e^{-\beta z^{i}} + A(1 - e^{-\beta z^{i}})$
\Else
\Statex \quad\ \textbf{Haze migration}
\State \quad\ $\delta = U(\delta_{min}, \delta_{max})$
\State \quad\ Uniformly sample an index $k$ $\sim$ $\{1,2,...,N\}$
\State \quad\ $ \textit{M}_{\delta} = \mathbbm{1}_{(h,w) \in [ -\delta H : \delta H, -\delta W : \delta W ]} $
\State \quad\ $I^{i}_{aug} =I^{k\rightarrow i} (I^{k},J^{i}) $
\EndIf
\State  \textbf{return} augmented sample $I_{aug}^{i}$
\end{algorithmic}
\label{algo}
\end{algorithm}

\textbf{Haze Migration.}
Haze effects in frequency domain, in contrast to boundary or shape, are commonly regarded as spatially static signals composed of low-frequency components (LFC). Yet the transferability of neural networks across LFC are not quite well. Frequency domain augmentation \cite{chen2021amplitude} is one way to alleviate this problem that distorts some frequency band/components such as LFC in images, and thereby alters their haze patterns. This is especially helpful in teaching networks about invariances to shifts in haze statistics. In particular, low-frequency spectrum of two images can be swapped without affecting the perception of high-level semantics \cite{chen2021amplitude, yang2020fda}. It inspires us to migrate only LFC carrying haze statistics from one image to another, which diversifies haze patterns but leaves scene understanding unchanged. Additionally, it breaks down the only position dependence of a hazy image $I$ on $d(p)$, resulting in more heterogeneous haze patterns. 

Consider two hazy images $I^a$ and $I^b$ synthesized from the groundtruth $J^a$ and $J^b$, and let $\mathcal{F}^A$, $\mathcal{F}^P$ be the amplitude and phase component of the Fourier transform  $\mathcal{F}$.
Similar to \cite{yang2020fda}, we denote an image mask $\textit{M}_{\delta}$ where only a small area at center is non-zero:
\begin{equation}
\textit{M}_{\delta}(h, w) = \mathbbm{1}_{(h,w) \in [ -\delta H : \delta H, -\delta W : \delta W ]}
  \label{eq:replace}
\end{equation}
here the image center is at $(0, 0)$ and the ratio $\delta \in (0, 1)$ controls the range of LFC to migrate.
Then the augmented image $I^{a\rightarrow b}$ can be formalized as:
\begin{scriptsize} 
\begin{equation}
  I^{a\rightarrow b} (I^{a},J^{b}) =  \mathcal{F}^{-1}([  \textit{M}_{\delta} \circ \mathcal{F}^A(I^{a}) + (1 - \textit{M}_{\delta}) \circ  \mathcal{F}^A(J^{b}), \mathcal{F}^P(J^{b}) ])
  \label{eq:fft}
\end{equation}
\end{scriptsize}where the range of LFC (\textit{i.e.,} haze pattern) of $J^{b}$ is replaced by that of $I^{a}$ and $\delta$ is carefully perturbed to preserve semantics in $J^{b}$. Note that unlike \cite{yang2020fda}, HazeAug is not for domain alignment but essentially an augmentation of hazing process within train data. At last, to mitigate the negative impacts of hard thresholding of $\delta$, a Gaussian filter is applied to smooth the image. 

The complete pipeline of HazeAug is summarized in Algo.\ref{algo}. We further demonstrate in Section \ref{sec:ablation} that diffusion models achieve significantly higher performance gain from the proposed augmentation strategy than other end-to-end methods.

\begin{table*}[!t]
	\centering
  \linespread{0.92}
 \small
	\caption{Quantitative comparison on four real-world datasets. \textbf{Bold} and \uline{underline} indicate the best and the second-best, respectively.}
 \resizebox{\textwidth}{!}{\begin{tabular}{c|c|cccccccccccc}
		\toprule
		Datasets & \multicolumn{1}{c|}{Indicators} & \multicolumn{1}{c}{DCP} & \multicolumn{1}{c}{Retinex} &  \multicolumn{1}{c}{UNet} & \multicolumn{1}{c}{FFANet} & \multicolumn{1}{c}{GCANet} & \multicolumn{1}{c}{DID} & \multicolumn{1}{c}{MSBDN} & \multicolumn{1}{c}{DF} & \multicolumn{1}{c}{$PSD_{\text{GCA}}$} & \multicolumn{1}{c}{$PSD_{\text{MSBDN}}$} & \multicolumn{1}{c}{D4} & \multicolumn{1}{c}{Ours} \\
		\midrule
		\multirow{5}[2]{*}{I-Haze} & PSNR $\uparrow$ & 16.037  & 15.088 &  16.485   & 16.887  & 16.979  & 16.565  & \uline{16.988}  & 16.550  & 14.694 & 14.822 & 16.938 & \textbf{17.346} \\
		& SSIM $\uparrow$   &  0.718  & 0.719  & 0.736  &  0.739  & 0.746  & 0.744  & 0.748  & \uline{0.754}  & 0.674 & 0.717 & 0.744 &\textbf{0.823} \\
		& CIEDE $\downarrow$    & 25.349  & 23.225 & 29.928 &  28.437  & 23.892  & 24.364  & 24.317  & \uline{12.567}  & 28.731  & 29.173 & 13.155 & \textbf{11.529} \\
            & MUSIQ $\uparrow$   & 58.964  & 61.488   & 45.617   & 46.755  & 62.118  & 63.101  & 63.340  & \uline{64.961}  & 58.399  & 63.011 & 64.514 & \textbf{68.792} \\
		& NIQE $\downarrow$    & 5.225  & 5.055  & 6.713  &  7.628  & 5.071  & 5.231  & 5.428  & 4.378  & 3.975 & \uline{3.912} & 4.539 & \textbf{3.817} \\
		\midrule
		\multirow{5}[2]{*}{O-Haze} & PSNR $\uparrow$   & 14.622  & 12.076   & 16.514    & 16.279  & 14.555  & 16.625 & 16.562  & 16.340  & 13.774 & 12.676 & \uline{16.922} & \textbf{17.389}  \\
		& SSIM $\uparrow$    & 0.676  & 0.594  & 0.657 &  0.661  & 0.652  & 0.656  & 0.651  & \uline{0.700}  & 0.567 & 0.605  & 0.607 & \textbf{0.802} \\
		& CIEDE $\downarrow$    & 25.956  & 27.922 & 23.814 & 23.952  & 25.069  & 23.551  & 23.284  & 15.897  & 28.480  & 24.031 & \uline{15.195} & \textbf{13.583} \\
            & MUSIQ $\uparrow$ & \uline{64.017}  & 61.535  & 62.232  & 62.700  & 63.388  & 63.269  & 63.157  & 63.908  & 56.650  & 63.047 & 63.042 & \textbf{64.549} \\
		& NIQE $\downarrow$ & 5.230  & 4.556 & 3.954  &  3.232  & 3.817  & 4.306  & 4.465  & 3.054  & 3.246 & 2.744 & \uline{2.610} & \textbf{2.268} \\
		\midrule
		\multirow{5}[2]{*}{Dense-Haze} & PSNR $\uparrow$  & 11.747  & \uline{12.278}   & 10.878 &  11.068  & 10.052  & 11.704  & 11.144  & 10.850  & 10.637 & 10.67 & 11.412 & \textbf{13.157} \\
		& SSIM $\uparrow$ & 0.416  & \uline{0.430}   & 0.389   & 0.395  & 0.387  & 0.403  & 0.392  & 0.381  & 0.386 & 0.399 & 0.385 & \textbf{0.487} \\
		& CIEDE $\downarrow$  & 29.821  & 25.683  & 25.541   & 25.267  & 27.519  & 24.358  & 24.935  & 25.524  & 30.159 & 29.470 & \uline{24.238} & \textbf{21.318} \\
            & MUSIQ $\uparrow$  & 43.092  & 41.603  & 41.346  & 42.068  & 41.620  & 42.852  & 41.645  & 42.083  & 35.311 & \uline{44.458} & 42.587 & \textbf{44.509} \\
		& NIQE $\downarrow$  & 6.804  & 6.886   & 6.455  &  5.905  & 6.703  & 5.820  & 6.843  & 6.391  & 5.988 & \uline{5.206} & 6.630 & \textbf{3.691} \\
		\midrule
		\multirow{5}[2]{*}{Nh-Haze} & PSNR $\uparrow$ & 12.489  & 10.075  &  12.329   & 12.246  & 11.632  & 12.346  & 12.500  & 12.250  & 11.356  & 12.135 & \uline{13.129} & \textbf{14.167} \\
		& SSIM $\uparrow$   & 0.514  & 0.430   & 0.484   & 0.481  & 0.493  & 0.477  & 0.477  & 0.498  & 0.465 & \uline{0.552} & 0.442 & \textbf{0.622} \\
		& CIEDE $\downarrow$  & 28.961  & 30.806   & 27.129   & 21.743  & 21.656  & 21.618  & 21.331  & 21.624  & 32.350  & 31.294 & \uline{20.830}  & \textbf{17.304} \\
            & MUSIQ $\uparrow$ & \uline{63.124}  & 60.953  & 53.331  & 61.634  & 62.199  & 62.312  & 61.965  & 62.999  & 58.546  & 61.196  & 60.991 & \textbf{63.675} \\
		& NIQE $\downarrow$   & 4.042  & 3.955  & 3.808  & 3.110   & 3.409  & 3.657  & 3.785  & 3.234  & 3.038  & 2.824 & \uline{2.680} & \textbf{2.383} \\
		\bottomrule
	\end{tabular}}%
	\label{Reference Metrics}%
\end{table*}%

\section{Experiments}
\label{sec:experiment}

In this section, extensive experiments are conducted to evaluate the proposed dehazing model on challenging real-world data. We firstly introduce detailed experiment settings and then compare to several state-of-the-art dehazing methods on multiple databases. It is followed by ablation studies on how the network design and data augmentation affect the behaviors of our dehazing model.

\subsection{Experimental settings}
\noindent \textbf{Databases.} We combine the Indoor Training Set (ITS) and Outdoor Training Set (OTS) of REalistic Single Image DEhazing (RESIDE) \cite{RESIDE} for model training. Five databases are utilized for performance evaluation: I-Haze \cite{IHAZE}, O-Haze \cite{OHAZE}, Dense-Haze \cite{densehaze}, Nh-Haze \cite{nhhaze} and RTTS \cite{RESIDE}.
All of them 
are captured in real environment, with Dense-Haze and Nh-Haze representing two challenging cases: heavy and non-homogeneous fog. Note that the first four databases have hazy-clean pairs, while RTTS only has hazy images. To emphasize the generalization ability to real and challenging scenarios, we only train the dehazing model on ITS and OTS and directly evaluate it on above five testsets without any fine-tune or re-train.

\noindent \textbf{Evaluation metrics.} To comprehensively evaluate the performances of various dehazing methods, reference and non-reference metrics are chosen to measure the fidelity and naturalness of dehazed images respectively. For reference metrics, PSNR\cite{PSNR1, PSNR2} , SSIM \cite{SSIM1, SSIM2} and CIEDE \cite{ciede2000-1, ciede2000-2} are used to measure the distortion between the dehazed and the groundtruth images from the aspect of pixel, structure and color respectively. For non-reference metrics, we use  NIQE \cite{NIQE} and MUSIQ\cite{MUSIQ} to measure how much the dehazed images are similar to natural images in terms of statistical regularities and granularities. 

\noindent \textbf{Compared methods.} 
We compare our model against current state-of-the-art dehazing algorithms based on conventional methods (DCP \cite{he2010single}, Retinex \cite{retinex}), convolutional neural networks (UNet \cite{unet}, FFANet \cite{FFA}, GCANet \cite{gcanet}, DID \cite{did}, MSBDN \cite{MSBDN}), and ViT \cite{dosovitskiy2020image} (DehazeFormer (DF) \cite{DehazeFormer}).
While they are all leading methods on synthetic data, we also compare to several state-of-the-art \textit{real} scene dehazing methods that require additional fine-tuning ($\text{PSD}_{\text{GCA}}$ \cite{PSD}, $\text{PSD}_{\text{MSBDN}}$ \cite{PSD} and D4 \cite{yang2022self}). For MSBDN, DF and $\text{PSD}_{\text{GCA}}$, we use the provided pre-trained models. For the rest, we retrain the models on the same dataset used in this paper using the provided codes for fair comparison.
For more implementation details, please see Appendix \textcolor{red}{A}.

\subsection{Performance Comparisons}

\begin{figure*}[!t]
	\centering
	\includegraphics [width=7in]{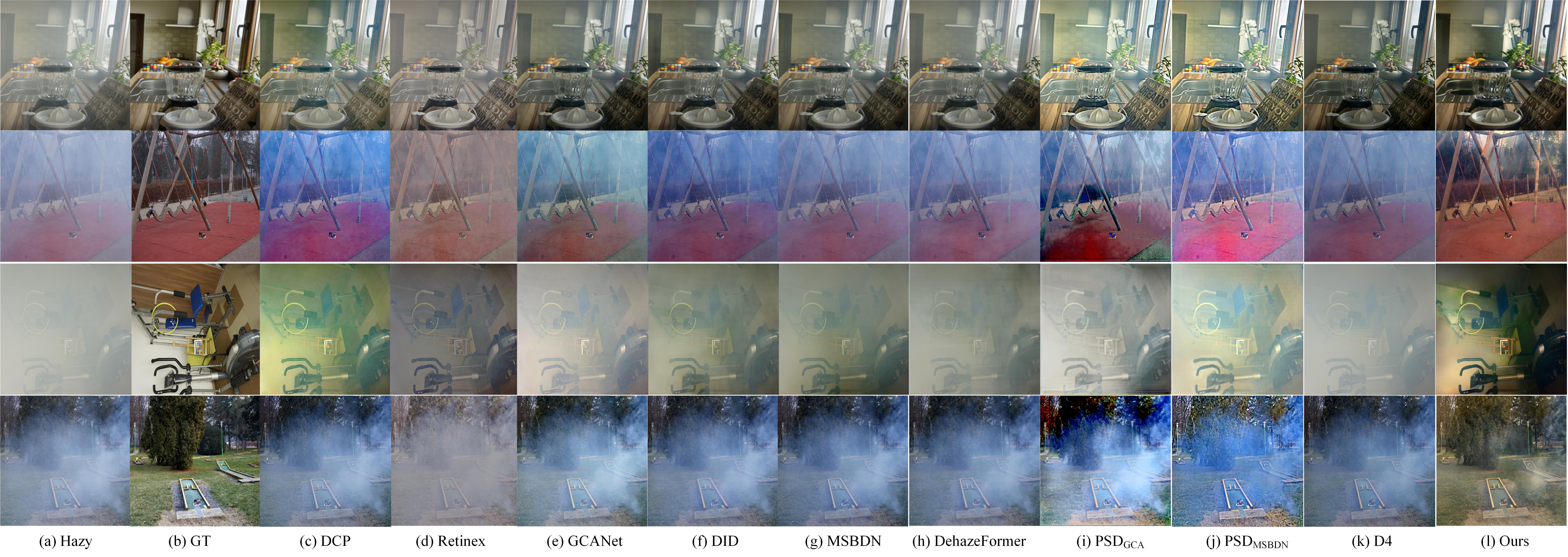}
	\caption{
	 Qualitative comparison of dehazed images with different methods.
  From top to bottom are the visualization on the I-Haze, O-Haze, Dense-Haze and Nh-Haze database. Additional results in the Appendix \textcolor{red}{C.1}.}
	\label{visual_sota}
\end{figure*}

\begin{table}[!t]
\centering
  \linespread{0.92}
 \scriptsize
 	\caption{Quantitative comparison on RTTS dataset. \textbf{Bold} and \underline{underline} indicate the best and the second-best, respectively.}
\resizebox{\linewidth}{!}{
\begin{tabular}{c|ccccccc}
\hline
Indicators & DCP & MSBDN       & DF          & $PSD_{\text{GCA}}$    & $PSD_{\text{MSBDN}}$  & D4    & Ours        \\ \hline
MUSIQ  $\uparrow$  & 55.079  & 54.559 & 55.884 & 53.401 & 56.712 & \underline{56.809} & \textbf{58.218} \\
NIQE  $\downarrow$  & 3.852   & 4.483 & 4.697 & 3.852 & \underline{3.667} & 4.415 & \textbf{2.968} \\ \hline
\end{tabular}
}
\label{RTTS}%
\end{table}


\noindent \textbf{Quantitative comparison.}
As showed in Table \ref{Reference Metrics},  our method largely outperforms other baselines across all the datasets. It is even superior to the real scene dehazing methods (PSD \cite{PSD} and D4 \cite{yang2022self}) that are fine-tuned on real hazy images. Another interesting observation is that conventional methods especially DCP are good at non-reference metrics but inferior on reference ones, demonstrating their emphasis on the naturalness of dehazed images. In contrast, deep learning based methods such as DF perform well on reference metrics because their learning objective (\textit{e.g.,} MSE) is to minimize distortion. Our method inherits the strengths of both diffusion models (which produce realistic images) and FCB (which recovers image details). As a result, we achieve state-of-the-art results on both fidelity and naturalness metrics.

Additionally, our method achieves significantly better performance on the challenging Dense-Haze and Nh-Haze databases. This is mainly attributed to HazeAug, which broadens the synthetic data distribution and teaches networks to be robust to extreme or heterogeneous hazy conditions. It is worth noting that when HazeAug is applied to other learning-based methods such as MSBDN (Appendix \textcolor{red}{B.4}), diffusion models still remain superior in performance. This further demonstrates the superiority of diffusion models in generalizing to real-world hazy images.

Table \ref{RTTS} shows the evaluation results on real-world hazy images in RTTS database. Substantial improvements on non-reference metrics, such as MUSIQ and NIQE, can be observed. It verifies effectiveness of our proposed model in real-world blind dehazing and its ability to produce dehazed images with high perceptual quality.

\begin{table}[t!]
\centering
 \scriptsize
\caption{Ablation study of FCB and HazeAug on real-world datasets. \textbf{Bold} and \uline{underline} indicate the best and the second-best, respectively.}
\resizebox{\linewidth}{!}{
\begin{tabular}{c|ccc|ccc}
\hline
\multirow{2}{*}{Datasets}   & \multicolumn{3}{c|}{Module}                & \multicolumn{3}{c}{Indicator}                     \\
 & DDPMs            & HazeAug                                                    & FCB                                                 & SSIM $\uparrow$ & CIEDE $\downarrow$ & MUSIQ $\uparrow$ \\ \hline
\multirow{4}{*}{I-Haze}                             & {\color[HTML]{333333} $\checkmark$} &                                                         &                                                                    & 0.536           & 18.671 & 60.730                 \\
                             & {\color[HTML]{333333} $\checkmark$} & \multicolumn{1}{c}{{\color[HTML]{333333} $\checkmark$}} &                                                     
            &          \uline{0.812} &         12.120 & \uline{68.353}        \\
                             & {\color[HTML]{333333} $\checkmark$} &                                                         & \multicolumn{1}{c|}{{\color[HTML]{333333} $\checkmark$}} &  0.747           &             \uline{12.047} & 65.571            \\
     & {\color[HTML]{333333} $\checkmark$} & \multicolumn{1}{c}{{\color[HTML]{333333} $\checkmark$}} & \multicolumn{1}{c|}{{\color[HTML]{333333} $\checkmark$}}         & \textbf{0.823}           & \textbf{11.529}  &  \textbf{68.792} \\ \hline
     \multirow{4}{*}{O-Haze}                        & {\color[HTML]{333333} $\checkmark$} &                                                         &                                                            & 0.708           & 18.006     & 62.241  \\
                             & {\color[HTML]{333333} $\checkmark$} & \multicolumn{1}{c}{{\color[HTML]{333333} $\checkmark$}} &                                                          &         \uline{0.776} &          \uline{15.081}   & \uline{64.219}  \\
                             & {\color[HTML]{333333} $\checkmark$} &                                                         & \multicolumn{1}{c|}{{\color[HTML]{333333} $\checkmark$}} & 0.712           &             15.805 & 63.463             \\
     & {\color[HTML]{333333} $\checkmark$} & \multicolumn{1}{c}{{\color[HTML]{333333} $\checkmark$}} & \multicolumn{1}{c|}{{\color[HTML]{333333} $\checkmark$}} &  \textbf{0.802}           & \textbf{13.583}    & \textbf{64.549}   \\ \hline
\end{tabular}
}
\label{module_aba}%
\end{table}

\noindent
\textbf{Qualitative comparison.}
Fig.\ref{visual_sota} compares the visual quality of dehazed images with different methods.
On I-haze and O-haze, our method is able to produce high-quality haze-free images with sharper edges and little haze residue. For dense and non-homogeneous haze, previous methods almost fail in dehazing, while our method is still able to recover most of the clean content despite some haze residue and occasional color distortion. Superior qualitative results are also seen on RTTS, which we provide in Appendix \textcolor{red}{C.1}.

\subsection{Ablation Study and Analysis}
\label{sec:ablation}


\begin{figure}[!t]
	\centering
	\includegraphics [width=3.0in]{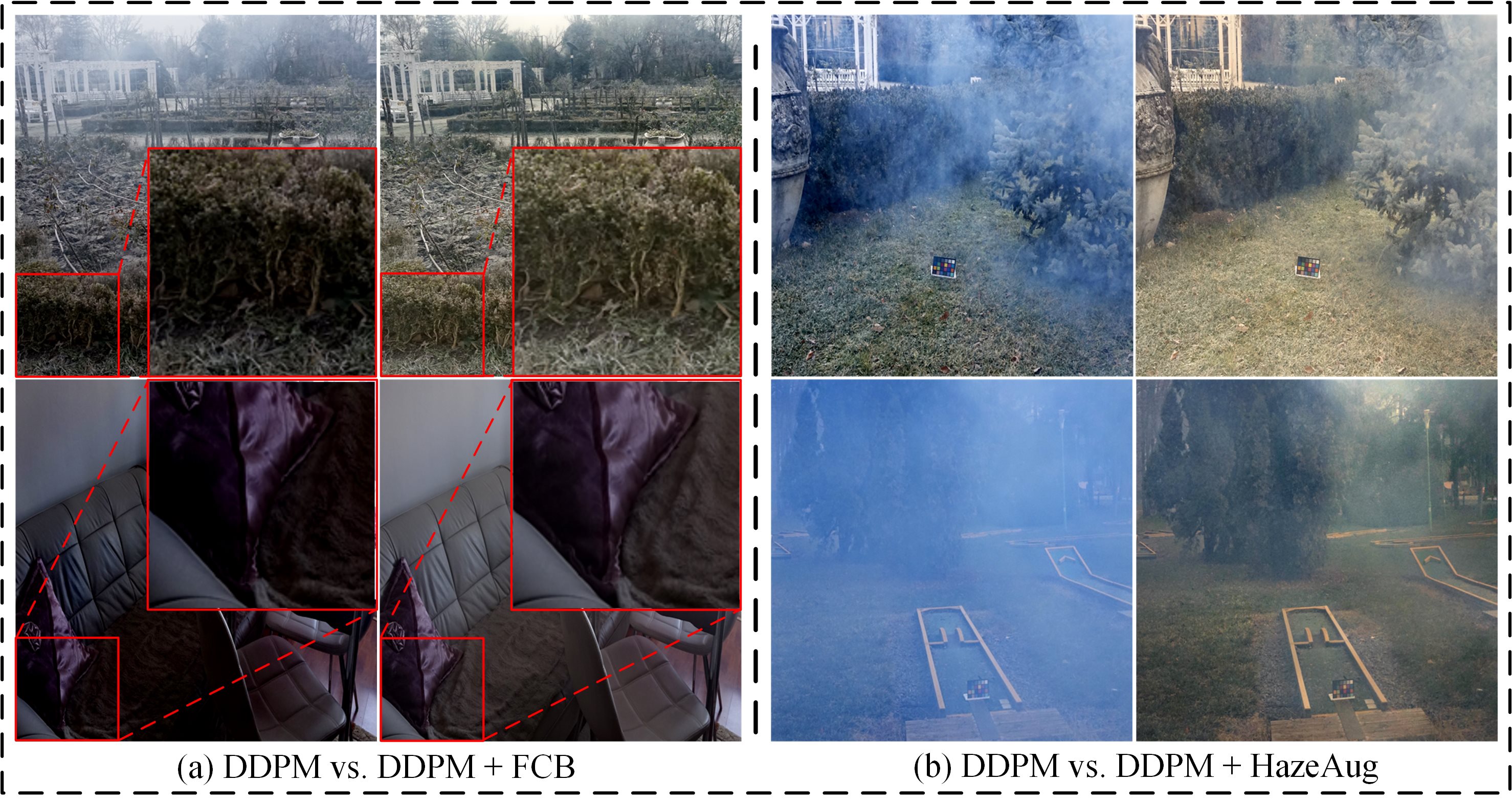}
	\caption{
	Representative dehazed images with ablation study. (a) DDPM vs. DDPM+FCB. (b) DDPM vs. DDPM+HazeAug.
	}
	\label{module_aba_visual}
\end{figure}

\noindent
\textbf{Ablation of two modules.}
We investigate the effectiveness of the proposed two modules and report the results in Table \ref{module_aba}.
A plain DDPM is actually outperformed by previous methods in terms of dehazing performance. Our modification to the network architecture with FCB makes a huge difference. In Table \ref{module_aba}, FCB significantly boosts the performance across all the metrics, including the reference-based ones, which are known to be challenging for diffusion models \cite{sr3, Whang_2022_CVPR, Palette}. This improvement can be visually observed in Fig.\ref{module_aba_visual}a, where the dehazed images from I-Haze and O-Haze are able to maintain finer image details and sharper edges. This further validates the role of FCB in recovering high-frequency information. Note that compared to Table \ref{Reference Metrics}, the proposed DDPM+FCB alone can match and even outperform the existing baselines in almost all the metrics.

As to HazeAug, it consistently improves all the metrics by a large margin in Table \ref{module_aba}. In Fig.\ref{module_aba_visual}b, we further show some difficult cases where HazeAug is able to remove most of the heavy haze, correct the color cast and enhance the contrast. These results demonstrate that effective haze augmentation can teach the model about invariances to the changing hazing process, which is crucial for improving generalization to real-world data.

Another interesting finding about HazeAug is that when applied to diffusion model and other learning-based methods (\textit{e.g.,} convolutional neural networks), diffusion models benefit more from the data augmentation (see Appendix \textcolor{red}{B.4}). It suggests the stronger ability of diffusion models in data modeling and their great potential in achieving high generality to real-world applications. 

\begin{figure}[!h]
	\centering
	\includegraphics [width=3.2in]{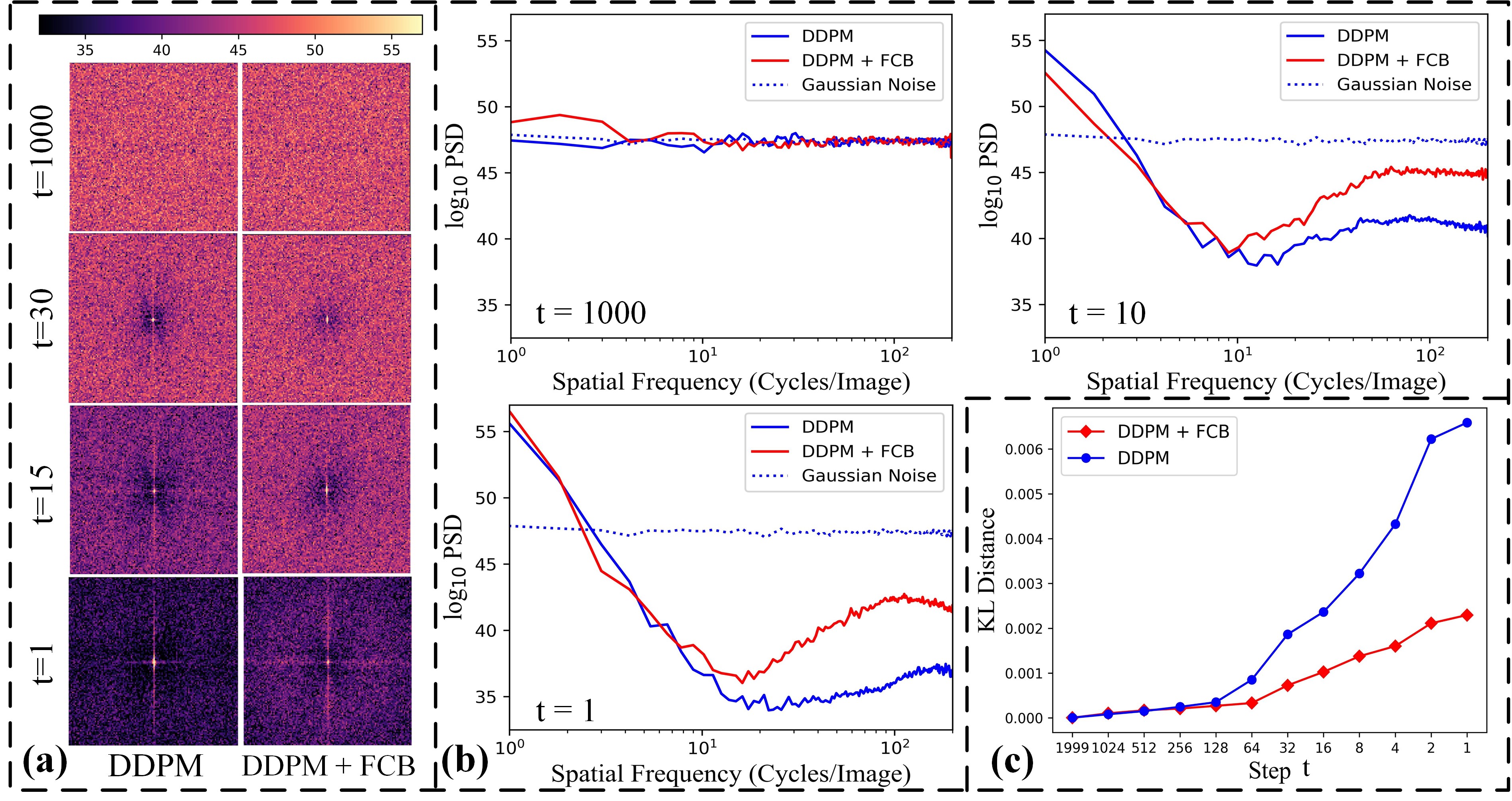}
	\caption{
Power spectrum analysis on $\epsilon$-prediction results of DDPMs at varying $t$.
(a) The power spectra of DDPM vs. DDPM+FCB.
(b) The PSD analysis of DDPM vs. DDPM+FCB.
(c) The KL distance between the spectrum of the predicted $\epsilon$ in (b) and that of the groundtruth. The smaller distance, the closer to groundtruth. Additional results are shown in the Appendix \textcolor{red}{B.2}.
	}
	\label{train_prove}
\end{figure}

\noindent
\textbf{Effect of frequency compensation.}
We examine how the proposed FCB helps with training.
Fig.\ref{train_prove}a visually compares the power spectrum \cite{sonka2014image} of the predicted $\epsilon$ between a plain DDPM and DDPM+FCB. when $t\rightarrow 1$, a considerably larger amount of higher frequency modes can be observed after introducing FCB.
This result is consistent with Fig.\ref{train_prove}b where the PSD analysis \footnote{pythonhosted.org/agpy} is performed on $\epsilon$-prediction. It validates that frequency compensation helps drive the output spectrum towards being uniformly distributed at smaller $t$.
In Fig.\ref{train_prove}c, we further use the KL distance \cite{murphy2012machine} to measure how far the spectrum of the predicted $\epsilon$ in Fig.\ref{train_prove}b deviates from the groundtruth Gaussian noise. We notice a clear advantage of using FCB to ease the learning process, especially at $t<100$.

\begin{table}[!h]
\scriptsize
\centering
\caption{Ablation study of FCB with varying number of Gaussian kernels. \textbf{Bold} and \uline{underline} indicate the best and the second-best.
}
\resizebox{\linewidth}{!}{
\begin{tabular}{c|c|c|ccc}
\hline
               Datasets        & k & $\sigma$  & 
               \multicolumn{1}{c}{SSIM $\uparrow$} & \multicolumn{1}{c}{CIEDE $\downarrow$} & \multicolumn{1}{c}{MUSIQ $\uparrow$} \\ \hline
\multirow{5}{*}{I-Haze}  & - &  - & 0.536 & 18.671                    & 60.730                 \\
                        &  \{ 7 \} &   \{ 10 \}  & 0.731    & 13.205    & 64.351                  \\
                        &  \{ 3,7 \} &  \{ 1,4 \}    & 0.732   & 12.909    & 65.372            \\
                        &  \{ 3,5,7 \} &  \{ 1,2,4 \} & \uline{0.747}       & \textbf{12.047} & {\uline{65.571}}                    \\
                        &  \{ 3,5,7,7 \} &  \{ 1,2,4,15 \}  & \textbf{0.754} & \uline{12.808}  & \textbf{65.714}            \\ \hline
\multirow{5}{*}{O-Haze} & - &  -    & 0.708  & 18.006                     & 62.241    \\
&  \{ 7 \} &   \{ 10 \} &  \textbf{0.714}    & 16.112    & 63.098          \\
&  \{ 3,7 \} &  \{ 1,4 \}    &  0.711   & 16.103    & \uline{63.192}       \\
&  \{ 3,5,7 \} &  \{ 1,2,4 \} & \uline{0.712}        & \textbf{15.805}  & \textbf{63.463}        \\
&  \{ 3,5,7,7 \} &  \{ 1,2,4,15 \} & 0.710 & \uline{15.998}  & 63.094  \\
\hline
\end{tabular}
}
\label{different_filters}
\end{table}

\noindent
\textbf{Ablation of filter design in FCB.}
We set up different filter configurations to identify the important factors of FCB design. We first vary the number of Gaussian kernels from $M=\{1, 2, 3, 4\}$ and choose $k$ and $\sigma$ in a way that different settings have similar bandwidth. Table \ref{different_filters} shows that even a single high-pass filter leads to a drastic improvement in performance. With more kernels, FCB can more flexibly adapt to training at varying $t$ and hence better performance. However, as the number of filters keeps increasing, the performance gain gradually saturates. We finally choose $M=3$ in this paper as it offers a well trade-off between the performance and computation efficiency. Additional ablation studies of filter design are shown in the Appendix \textcolor{red}{B.3}.

\begin{table}[h!]
\centering
\scriptsize
\caption{Ablation study of HazeAug.  \textbf{Bold} and \uline{underline} indicate the best and the second-best, respectively.}
\resizebox{\linewidth}{!}{
\begin{tabular}{c|ccc|ccccc}
\hline
\multirow{2}{*}{Datasets}   & \multicolumn{3}{c|}{Augmentation Methods}                & \multicolumn{3}{c}{Indicator}                     \\ 
                            & Traditional  & Hard Sample       & Haze Migration     & SSIM $\uparrow$ & CIEDE $\downarrow$ & MUSIQ $\uparrow$  \\ \hline
\multirow{4}{*}{I-Haze}     & $\checkmark$ &              &                & 0.536   & 18.671   & 60.730     \\
                            &              & $\checkmark$ &                & \uline{0.810}   &    \uline{14.647}      & \uline{68.341}     \\
                            &              &              & $\checkmark$   &     0.745    &   15.341       &    66.164       \\
                            &              & $\checkmark$ & $\checkmark$ 
            &          \textbf{0.812} &         \textbf{12.120}           &       \textbf{68.353}         \\ \hline
\multirow{4}{*}{O-Haze}     & $\checkmark$ &              &                & 0.708   & 18.006   & 62.241      \\
                            &              & $\checkmark$ &                & \uline{0.750}   &     \uline{16.314}     & \textbf{66.383}      \\
                            &              &              & $\checkmark$ &      0.732      &     16.487     &   62.341           \\
                            &              & $\checkmark$ & $\checkmark$ 
           &         \textbf{0.776} &          \textbf{15.081} &           \uline{64.219}         \\ \hline
\end{tabular}
}
\label{hsv aba}
\end{table}

\noindent
\textbf{Ablation of HazeAug.}
In Table \ref{hsv aba}, we study the two operators in HazeAug, \textit{hard sample} and \textit{haze migration}.
Interestingly, although designed to handle severe hazy conditions, such as Dense- and Nh-haze, \textit{hard sample} also drastically improves the performance on I-Haze and O-Haze. A possible reason is that shielding the clean content from the diffusion models by severe haze forces them to rely less on its visual cues, but encourages them to do more generative modeling.

As for \textit{haze migration}, we observe considerable improvements across all metrics, especially the reference-based ones. Such distortion reduction can be attributed to the LFC migration, which implicitly forces the model to learn robust high-frequency signals that are invariant to the changing low-frequency ones.

\section{Conclusion}
\label{sec:conclusion}

We propose a dehazing framework based on conditional diffusion models which demonstrates superior generalization ability to real haze via comprehensive experiments. Specifically, we design a novel network unit, FCB, to ease the learning and provide insights into the limitation of diffusion models in restoring high-frequency image details. We hope it may prove useful for other inverse problems. Then, we find that diffusion models significantly benefit from effectively augmented data for improved generality. It shows potential of diffusion models to handle challenging issues (\textit{e.g.,} heterogeneous haze, color distortion) if applied with more sophisticated haze augmentation.

\section{Acknowledgement}
\label{sec:acknowledgement}
All the work is done at Sony R \& D Center China, Beijing Lab, supported by Sony Research Fund from June 2022 to May 2023. 


{\small
 \bibliographystyle{ieee_fullname}
 \bibliography{egbib}

\begin{thebibliography}{10}\itemsep=-1pt

\bibitem{IHAZE}
Cosmin Ancuti, Codruta~O Ancuti, Radu Timofte, and Christophe~De Vleeschouwer.
\newblock I-haze: a dehazing benchmark with real hazy and haze-free indoor
  images.
\newblock In {\em International Conference on Advanced Concepts for Intelligent
  Vision Systems}, pages 620--631. Springer, 2018.

\bibitem{densehaze}
Codruta~O Ancuti, Cosmin Ancuti, Mateu Sbert, and Radu Timofte.
\newblock Dense-haze: A benchmark for image dehazing with dense-haze and
  haze-free images.
\newblock In {\em 2019 IEEE international conference on image processing
  (ICIP)}, pages 1014--1018. IEEE, 2019.

\bibitem{nhhaze}
Codruta~O Ancuti, Cosmin Ancuti, and Radu Timofte.
\newblock Nh-haze: An image dehazing benchmark with non-homogeneous hazy and
  haze-free images.
\newblock In {\em Proceedings of the IEEE/CVF conference on computer vision and
  pattern recognition workshops}, pages 444--445, 2020.

\bibitem{OHAZE}
Codruta~O Ancuti, Cosmin Ancuti, Radu Timofte, and Christophe De~Vleeschouwer.
\newblock O-haze: a dehazing benchmark with real hazy and haze-free outdoor
  images.
\newblock In {\em Proceedings of the IEEE conference on computer vision and
  pattern recognition workshops}, pages 754--762, 2018.

\bibitem{avrahami2022blended}
Omri Avrahami, Dani Lischinski, and Ohad Fried.
\newblock Blended diffusion for text-driven editing of natural images.
\newblock In {\em Proceedings of the IEEE/CVF Conference on Computer Vision and
  Pattern Recognition}, pages 18208--18218, 2022.

\bibitem{tip2022}
Haoran Bai, Jinshan Pan, Xinguang Xiang, and Jinhui Tang.
\newblock Self-guided image dehazing using progressive feature fusion.
\newblock {\em IEEE Transactions on Image Processing}, 31:1217--1229, 2022.

\bibitem{13}
Dana Berman, Shai Avidan, et~al.
\newblock Non-local image dehazing.
\newblock In {\em Proceedings of the IEEE conference on computer vision and
  pattern recognition}, pages 1674--1682, 2016.

\bibitem{boreman2001modulation}
Glenn~D Boreman.
\newblock {\em Modulation transfer function in optical and electro-optical
  systems}, volume~4.
\newblock SPIE press Bellingham, Washington, 2001.

\bibitem{bui2017single}
Trung~Minh Bui and Wonha Kim.
\newblock Single image dehazing using color ellipsoid prior.
\newblock {\em IEEE Transactions on Image Processing}, 27(2):999--1009, 2017.

\bibitem{bundy1984difference}
Alan Bundy and Lincoln Wallen.
\newblock Difference of gaussians.
\newblock {\em Catalogue of Artificial Intelligence Tools}, pages 30--30, 1984.

\bibitem{14}
Bolun Cai, Xiangmin Xu, Kui Jia, Chunmei Qing, and Dacheng Tao.
\newblock Dehazenet: An end-to-end system for single image haze removal.
\newblock {\em IEEE Transactions on Image Processing}, 25(11):5187--5198, 2016.

\bibitem{gcanet}
Dongdong Chen, Mingming He, Qingnan Fan, Jing Liao, Liheng Zhang, Dongdong Hou,
  Lu Yuan, and Gang Hua.
\newblock Gated context aggregation network for image dehazing and deraining.
\newblock In {\em 2019 IEEE winter conference on applications of computer
  vision (WACV)}, pages 1375--1383. IEEE, 2019.

\bibitem{chen2021amplitude}
Guangyao Chen, Peixi Peng, Li Ma, Jia Li, Lin Du, and Yonghong Tian.
\newblock Amplitude-phase recombination: Rethinking robustness of convolutional
  neural networks in frequency domain.
\newblock In {\em Proceedings of the IEEE/CVF International Conference on
  Computer Vision}, pages 458--467, 2021.

\bibitem{PSD}
Zeyuan Chen, Yangchao Wang, Yang Yang, and Dong Liu.
\newblock Psd: Principled synthetic-to-real dehazing guided by physical priors.
\newblock In {\em Proceedings of the IEEE/CVF conference on computer vision and
  pattern recognition}, pages 7180--7189, 2021.

\bibitem{dhariwal2021diffusion}
Prafulla Dhariwal and Alexander Nichol.
\newblock Diffusion models beat gans on image synthesis.
\newblock {\em Advances in Neural Information Processing Systems},
  34:8780--8794, 2021.

\bibitem{MSBDN}
Hang Dong, Jinshan Pan, Lei Xiang, Zhe Hu, Xinyi Zhang, Fei Wang, and
  Ming-Hsuan Yang.
\newblock Multi-scale boosted dehazing network with dense feature fusion.
\newblock In {\em Proceedings of the IEEE/CVF conference on computer vision and
  pattern recognition}, pages 2157--2167, 2020.

\bibitem{dong2020fd}
Yu Dong, Yihao Liu, He Zhang, Shifeng Chen, and Yu Qiao.
\newblock Fd-gan: Generative adversarial networks with fusion-discriminator for
  single image dehazing.
\newblock In {\em Proceedings of the AAAI Conference on Artificial
  Intelligence}, volume~34, pages 10729--10736, 2020.

\bibitem{dosovitskiy2020image}
Alexey Dosovitskiy, Lucas Beyer, Alexander Kolesnikov, Dirk Weissenborn,
  Xiaohua Zhai, Thomas Unterthiner, Mostafa Dehghani, Matthias Minderer, Georg
  Heigold, Sylvain Gelly, et~al.
\newblock An image is worth 16x16 words: Transformers for image recognition at
  scale.
\newblock {\em arXiv preprint arXiv:2010.11929}, 2020.

\bibitem{engin2018cycle}
Deniz Engin, Anil Gen{\c{c}}, and Hazim Kemal~Ekenel.
\newblock Cycle-dehaze: Enhanced cyclegan for single image dehazing.
\newblock In {\em Proceedings of the IEEE conference on computer vision and
  pattern recognition workshops}, pages 825--833, 2018.

\bibitem{fattal2014dehazing}
Raanan Fattal.
\newblock Dehazing using color-lines.
\newblock {\em ACM transactions on graphics (TOG)}, 34(1):1--14, 2014.

\bibitem{monodepth}
Cl{\'{e}}ment Godard, Oisin {Mac Aodha}, and Gabriel~J. Brostow.
\newblock Unsupervised monocular depth estimation with left-right consistency.
\newblock In {\em CVPR}, 2017.

\bibitem{goodfellow2020generative}
Ian Goodfellow, Jean Pouget-Abadie, Mehdi Mirza, Bing Xu, David Warde-Farley,
  Sherjil Ozair, Aaron Courville, and Yoshua Bengio.
\newblock Generative adversarial networks.
\newblock {\em Communications of the ACM}, 63(11):139--144, 2020.

\bibitem{20}
Kaiming He, Xinlei Chen, Saining Xie, Yanghao Li, Piotr Doll{\'a}r, and Ross
  Girshick.
\newblock Masked autoencoders are scalable vision learners.
\newblock In {\em Proceedings of the IEEE/CVF Conference on Computer Vision and
  Pattern Recognition}, pages 16000--16009, 2022.

\bibitem{he2010single}
Kaiming He, Jian Sun, and Xiaoou Tang.
\newblock Single image haze removal using dark channel prior.
\newblock {\em IEEE transactions on pattern analysis and machine intelligence},
  33(12):2341--2353, 2010.

\bibitem{ho2020denoising}
Jonathan Ho, Ajay Jain, and Pieter Abbeel.
\newblock Denoising diffusion probabilistic models.
\newblock {\em Advances in Neural Information Processing Systems},
  33:6840--6851, 2020.

\bibitem{ho2022classifier}
Jonathan Ho and Tim Salimans.
\newblock Classifier-free diffusion guidance.
\newblock {\em arXiv preprint arXiv:2207.12598}, 2022.

\bibitem{PSNR1}
Alain Hore and Djemel Ziou.
\newblock Image quality metrics: Psnr vs. ssim.
\newblock In {\em 2010 20th international conference on pattern recognition},
  pages 2366--2369. IEEE, 2010.

\bibitem{jun2020distribution}
Heewoo Jun, Rewon Child, Mark Chen, John Schulman, Aditya Ramesh, Alec Radford,
  and Ilya Sutskever.
\newblock Distribution augmentation for generative modeling.
\newblock In {\em International Conference on Machine Learning}, pages
  5006--5019. PMLR, 2020.

\bibitem{MUSIQ}
Junjie Ke, Qifei Wang, Yilin Wang, Peyman Milanfar, and Feng Yang.
\newblock Musiq: Multi-scale image quality transformer.
\newblock In {\em Proceedings of the IEEE/CVF International Conference on
  Computer Vision}, pages 5148--5157, 2021.

\bibitem{li2017aod}
Boyi Li, Xiulian Peng, Zhangyang Wang, Jizheng Xu, and Dan Feng.
\newblock Aod-net: All-in-one dehazing network.
\newblock In {\em Proceedings of the IEEE international conference on computer
  vision}, pages 4770--4778, 2017.

\bibitem{RESIDE}
Boyi Li, Wenqi Ren, Dengpan Fu, Dacheng Tao, Dan Feng, Wenjun Zeng, and
  Zhangyang Wang.
\newblock Benchmarking single-image dehazing and beyond.
\newblock {\em IEEE Transactions on Image Processing}, 28(1):492--505, 2018.

\bibitem{li2019pdr}
Chongyi Li, Chunle Guo, Jichang Guo, Ping Han, Huazhu Fu, and Runmin Cong.
\newblock Pdr-net: Perception-inspired single image dehazing network with
  refinement.
\newblock {\em IEEE Transactions on Multimedia}, 22(3):704--716, 2019.

\bibitem{li2017dr}
Chongyi Li, Jichang Guo, Fatih Porikli, Chunle Guo, Huzhu Fu, and Xi Li.
\newblock Dr-net: transmission steered single image dehazing network with
  weakly supervised refinement.
\newblock {\em arXiv preprint arXiv:1712.00621}, 2017.

\bibitem{liu2020end}
Wei Liu, Xianxu Hou, Jiang Duan, and Guoping Qiu.
\newblock End-to-end single image fog removal using enhanced cycle consistent
  adversarial networks.
\newblock {\em IEEE Transactions on Image Processing}, 29:7819--7833, 2020.

\bibitem{liu2019learning}
Yang Liu, Jinshan Pan, Jimmy Ren, and Zhixun Su.
\newblock Learning deep priors for image dehazing.
\newblock In {\em Proceedings of the IEEE/CVF international conference on
  computer vision}, pages 2492--2500, 2019.

\bibitem{did}
Ye Liu, Lei Zhu, Shunda Pei, Huazhu Fu, Jing Qin, Qing Zhang, Liang Wan, and
  Wei Feng.
\newblock From synthetic to real: Image dehazing collaborating with unlabeled
  real data.
\newblock In {\em Proceedings of the 29th ACM International Conference on
  Multimedia}, pages 50--58, 2021.

\bibitem{Sdedit}
Chenlin Meng, Yang Song, Jiaming Song, Jiajun Wu, Jun-Yan Zhu, and Stefano
  Ermon.
\newblock Sdedit: Image synthesis and editing with stochastic differential
  equations.
\newblock {\em arXiv preprint arXiv:2108.01073}, 2021.

\bibitem{murphy2012machine}
Kevin~P Murphy.
\newblock {\em Machine learning: a probabilistic perspective}.
\newblock MIT press, 2012.

\bibitem{nayar1999vision}
Shree~K Nayar and Srinivasa~G Narasimhan.
\newblock Vision in bad weather.
\newblock In {\em Proceedings of the seventh IEEE international conference on
  computer vision}, volume~2, pages 820--827. IEEE, 1999.

\bibitem{nishino2012bayesian}
Ko Nishino, Louis Kratz, and Stephen Lombardi.
\newblock Bayesian defogging.
\newblock {\em International journal of computer vision}, 98:263--278, 2012.

\bibitem{FFA}
Xu Qin, Zhilin Wang, Yuanchao Bai, Xiaodong Xie, and Huizhu Jia.
\newblock Ffa-net: Feature fusion attention network for single image dehazing.
\newblock In {\em Proceedings of the AAAI Conference on Artificial
  Intelligence}, volume~34, pages 11908--11915, 2020.

\bibitem{rahaman2019spectral}
Nasim Rahaman, Aristide Baratin, Devansh Arpit, Felix Draxler, Min Lin, Fred
  Hamprecht, Yoshua Bengio, and Aaron Courville.
\newblock On the spectral bias of neural networks.
\newblock In {\em International Conference on Machine Learning}, pages
  5301--5310. PMLR, 2019.

\bibitem{retinex}
Zia-ur Rahman, Daniel~J Jobson, and Glenn~A Woodell.
\newblock Multi-scale retinex for color image enhancement.
\newblock In {\em Proceedings of 3rd IEEE international conference on image
  processing}, volume~3, pages 1003--1006. IEEE, 1996.

\bibitem{rainal1961sampling}
AJ Rainal.
\newblock Sampling technique for generating gaussian noise.
\newblock {\em Review of Scientific Instruments}, 32(3):327--331, 1961.

\bibitem{unet}
Olaf Ronneberger, Philipp Fischer, and Thomas Brox.
\newblock U-net: Convolutional networks for biomedical image segmentation.
\newblock In {\em International Conference on Medical image computing and
  computer-assisted intervention}, pages 234--241. Springer, 2015.

\bibitem{Palette}
Chitwan Saharia, William Chan, Huiwen Chang, Chris Lee, Jonathan Ho, Tim
  Salimans, David Fleet, and Mohammad Norouzi.
\newblock Palette: Image-to-image diffusion models.
\newblock In {\em ACM SIGGRAPH 2022 Conference Proceedings}, pages 1--10, 2022.

\bibitem{saharia2021image}
Chitwan Saharia, Jonathan Ho, William Chan, Tim Salimans, David~J Fleet, and
  Mohammad Norouzi.
\newblock Image super-resolution via iterative refinement.
\newblock {\em arXiv preprint arXiv:2104.07636}, 2021.

\bibitem{sr3}
Chitwan Saharia, Jonathan Ho, William Chan, Tim Salimans, David~J Fleet, and
  Mohammad Norouzi.
\newblock Image super-resolution via iterative refinement.
\newblock {\em IEEE Transactions on Pattern Analysis and Machine Intelligence},
  2022.

\bibitem{shao2020domain}
Yuanjie Shao, Lerenhan Li, Wenqi Ren, Changxin Gao, and Nong Sang.
\newblock Domain adaptation for image dehazing.
\newblock In {\em Proceedings of the IEEE/CVF conference on computer vision and
  pattern recognition}, pages 2808--2817, 2020.

\bibitem{ciede2000-1}
Gaurav Sharma, Wencheng Wu, and Edul~N Dalal.
\newblock The ciede2000 color-difference formula: Implementation notes,
  supplementary test data, and mathematical observations.
\newblock {\em Color Research \& Application: Endorsed by Inter-Society Color
  Council, The Colour Group (Great Britain), Canadian Society for Color, Color
  Science Association of Japan, Dutch Society for the Study of Color, The
  Swedish Colour Centre Foundation, Colour Society of Australia, Centre
  Fran{\c{c}}ais de la Couleur}, 30(1):21--30, 2005.

\bibitem{NIQE}
Hamid~R Sheikh, Alan~C Bovik, and Lawrence Cormack.
\newblock No-reference quality assessment using natural scene statistics:
  Jpeg2000.
\newblock {\em IEEE Transactions on image processing}, 14(11):1918--1927, 2005.

\bibitem{DDPM-theory}
Jascha Sohl-Dickstein, Eric Weiss, Niru Maheswaranathan, and Surya Ganguli.
\newblock Deep unsupervised learning using nonequilibrium thermodynamics.
\newblock In {\em International Conference on Machine Learning}, pages
  2256--2265. PMLR, 2015.

\bibitem{ddim}
Jiaming Song, Chenlin Meng, and Stefano Ermon.
\newblock Denoising diffusion implicit models.
\newblock {\em arXiv preprint arXiv:2010.02502}, 2020.

\bibitem{DDPM-song}
Yang Song and Stefano Ermon.
\newblock Generative modeling by estimating gradients of the data distribution.
\newblock {\em Advances in Neural Information Processing Systems}, 32, 2019.

\bibitem{DehazeFormer}
Yuda Song, Zhuqing He, Hui Qian, and Xin Du.
\newblock Vision transformers for single image dehazing.
\newblock {\em arXiv preprint arXiv:2204.03883}, 2022.

\bibitem{song2021solving}
Yang Song, Liyue Shen, Lei Xing, and Stefano Ermon.
\newblock Solving inverse problems in medical imaging with score-based
  generative models.
\newblock {\em arXiv preprint arXiv:2111.08005}, 2021.

\bibitem{sonka2014image}
Milan Sonka, Vaclav Hlavac, and Roger Boyle.
\newblock {\em Image processing, analysis, and machine vision}.
\newblock Cengage Learning, 2014.

\bibitem{PSNR2}
Alexander Tanchenko.
\newblock Visual-psnr measure of image quality.
\newblock {\em Journal of Visual Communication and Image Representation},
  25(5):874--878, 2014.

\bibitem{tang2014investigating}
Ketan Tang, Jianchao Yang, and Jue Wang.
\newblock Investigating haze-relevant features in a learning framework for
  image dehazing.
\newblock In {\em Proceedings of the IEEE conference on computer vision and
  pattern recognition}, pages 2995--3000, 2014.

\bibitem{van1996modelling}
van~A Van~der Schaaf and JH~van van Hateren.
\newblock Modelling the power spectra of natural images: statistics and
  information.
\newblock {\em Vision research}, 36(17):2759--2770, 1996.

\bibitem{wang2021fully}
Cong Wang, Yan Huang, Yuexian Zou, and Yong Xu.
\newblock Fully non-homogeneous atmospheric scattering modeling with
  convolutional neural networks for single image dehazing.
\newblock {\em arXiv preprint arXiv:2108.11292}, 2021.

\bibitem{wang2021fwb}
Cong Wang, Yan Huang, Yuexian Zou, and Yong Xu.
\newblock Fwb-net: front white balance network for color shift correction in
  single image dehazing via atmospheric light estimation.
\newblock In {\em ICASSP 2021-2021 IEEE International Conference on Acoustics,
  Speech and Signal Processing (ICASSP)}, pages 2040--2044. IEEE, 2021.

\bibitem{wang2020high}
Haohan Wang, Xindi Wu, Zeyi Huang, and Eric~P Xing.
\newblock High-frequency component helps explain the generalization of
  convolutional neural networks.
\newblock In {\em Proceedings of the IEEE/CVF Conference on Computer Vision and
  Pattern Recognition}, pages 8684--8694, 2020.

\bibitem{SSIM2}
Zhou Wang, Alan~C Bovik, Hamid~R Sheikh, and Eero~P Simoncelli.
\newblock Image quality assessment: from error visibility to structural
  similarity.
\newblock {\em IEEE transactions on image processing}, 13(4):600--612, 2004.

\bibitem{SSIM1}
Zhou Wang, Eero~P Simoncelli, and Alan~C Bovik.
\newblock Multiscale structural similarity for image quality assessment.
\newblock In {\em The Thrity-Seventh Asilomar Conference on Signals, Systems \&
  Computers, 2003}, volume~2, pages 1398--1402. Ieee, 2003.

\bibitem{ciede2000-2}
Stephen Westland, Caterina Ripamonti, and Vien Cheung.
\newblock {\em Computational colour science using MATLAB}.
\newblock John Wiley \& Sons, 2012.

\bibitem{Whang_2022_CVPR}
Jay Whang, Mauricio Delbracio, Hossein Talebi, Chitwan Saharia, Alexandros~G.
  Dimakis, and Peyman Milanfar.
\newblock Deblurring via stochastic refinement.
\newblock In {\em Proceedings of the IEEE/CVF Conference on Computer Vision and
  Pattern Recognition (CVPR)}, pages 16293--16303, June 2022.

\bibitem{xu2012fast}
Haoran Xu, Jianming Guo, Qing Liu, and Lingli Ye.
\newblock Fast image dehazing using improved dark channel prior.
\newblock In {\em 2012 IEEE international conference on information science and
  technology}, pages 663--667. IEEE, 2012.

\bibitem{yang2020fda}
Yanchao Yang and Stefano Soatto.
\newblock Fda: Fourier domain adaptation for semantic segmentation.
\newblock In {\em Proceedings of the IEEE/CVF Conference on Computer Vision and
  Pattern Recognition}, pages 4085--4095, 2020.

\bibitem{zhang2018densely}
He Zhang and Vishal~M Patel.
\newblock Densely connected pyramid dehazing network.
\newblock In {\em Proceedings of the IEEE conference on computer vision and
  pattern recognition}, pages 3194--3203, 2018.

\bibitem{zhao2017deep}
Xi Zhao, Keyan Wang, Yunsong Li, and Jiaojiao Li.
\newblock Deep fully convolutional regression networks for single image haze
  removal.
\newblock In {\em 2017 IEEE Visual Communications and Image Processing (VCIP)},
  pages 1--4. IEEE, 2017.

\bibitem{zhu2017unpaired}
Jun-Yan Zhu, Taesung Park, Phillip Isola, and Alexei~A Efros.
\newblock Unpaired image-to-image translation using cycle-consistent
  adversarial networks.
\newblock In {\em Proceedings of the IEEE international conference on computer
  vision}, pages 2223--2232, 2017.

\bibitem{zhu2015fast}
Qingsong Zhu, Jiaming Mai, and Ling Shao.
\newblock A fast single image haze removal algorithm using color attenuation
  prior.
\newblock {\em IEEE transactions on image processing}, 24(11):3522--3533, 2015.

\end{thebibliography}
}


\end{document}


\title{Supplemental Material}
\maketitle

\appendix
\label{sec:appendix}

\section{Implementation Details}
\noindent
\textbf{Architecture and augmentation details.}
We adopt the UNET architecture of SR3 [48] as the backbone,
and replace all of its skip connections to FCBs.
Note that our network contains a total of 10 FCBs due to the $\frac{1}{\sqrt{2}}$ upsampling scale of skip connections in SR3.
Within FCB, we initialize the trainable weights to 1, and set $k \in \{3, 5, 7\}$ and $\sigma \in \{1, 2, 4\}$ by default.
As to HazeAug, MonoDepth [21] is utilized as the depth predictor.
Compared to the settings in [31], we adjust $(A_{min}, A_{max})$ from $(0.7, 1.0)$ to (0.5, 1.8) and $(\beta_{min}, \beta_{max})$ from (0.6, 1.8) to (0.8, 2.8). Finally, $(\delta_{min}, \delta_{max})$ are set to (0, 2.1e$^{-3}$).

\noindent
\textbf{Training details.}
During training, we resize each image to 128$\times$128 with the batch size of 3. We use step $T=2000$ for all experiments in the training. 
The learning rate is set to 1e-4 with a decay of 0.7 every 0.4M iterations and the models are trained for 2M iterations in total. 

\noindent
\textbf{Evaluation details.}
During testing, we follow [6] to resize the input image to 512$\times$512 for all the models to be evaluated.
To accelerate the sampling process, DDIM [53]is utilized with 20 steps to generate a dehazed image whose performance is similar to DDPM (see Appendix \ref{DDIM_sample}).
Despite the non-deterministic outputs of DDIM with respect to varying initial Gaussian noises, we only observe small variances among the outputs and decide to average on five sampling results to get the final one.

\section{Additional Ablation Results}

\subsection{Sampling strategy of conditional DDPM}
\label{DDIM_sample}

\begin{figure}[!t]
	\centering
	\includegraphics [width=3.2in]{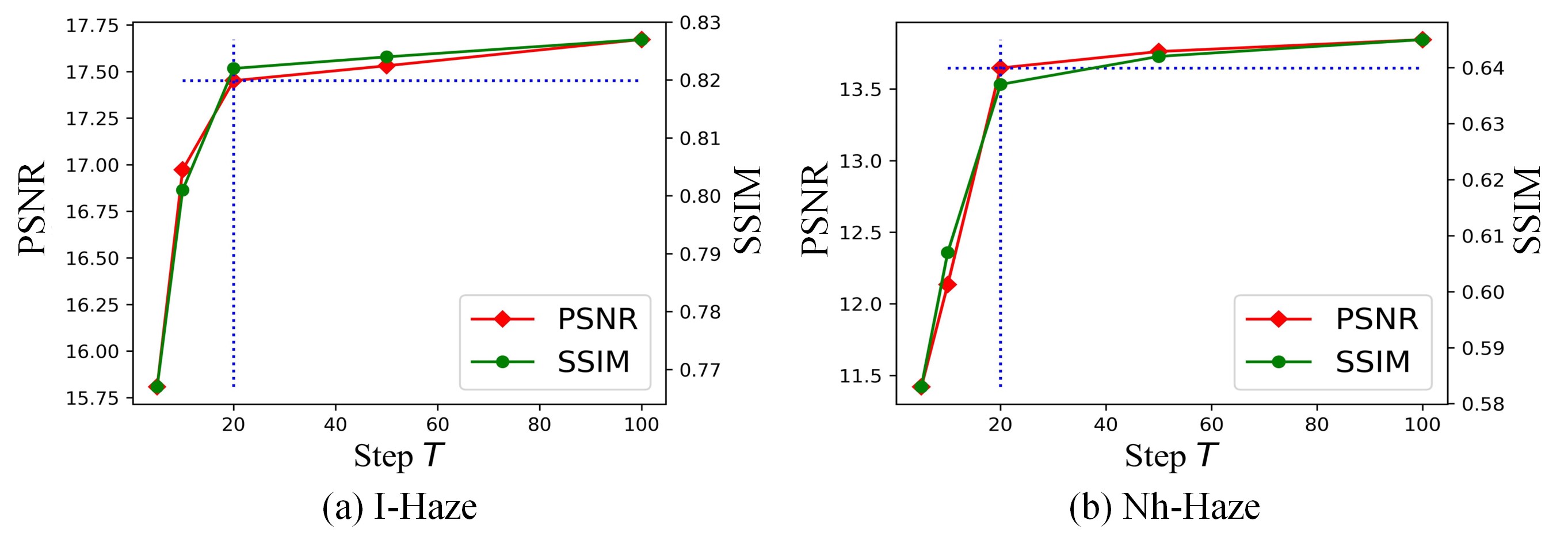}
	\caption{
	Influence of the sampling step T on PSNR and SSIM with the DDIM sampling procedure.
 }
	\label{diff_dimm_aba}
\end{figure}

\begin{figure}[!t]
	\centering
	\includegraphics [width=3.2in]{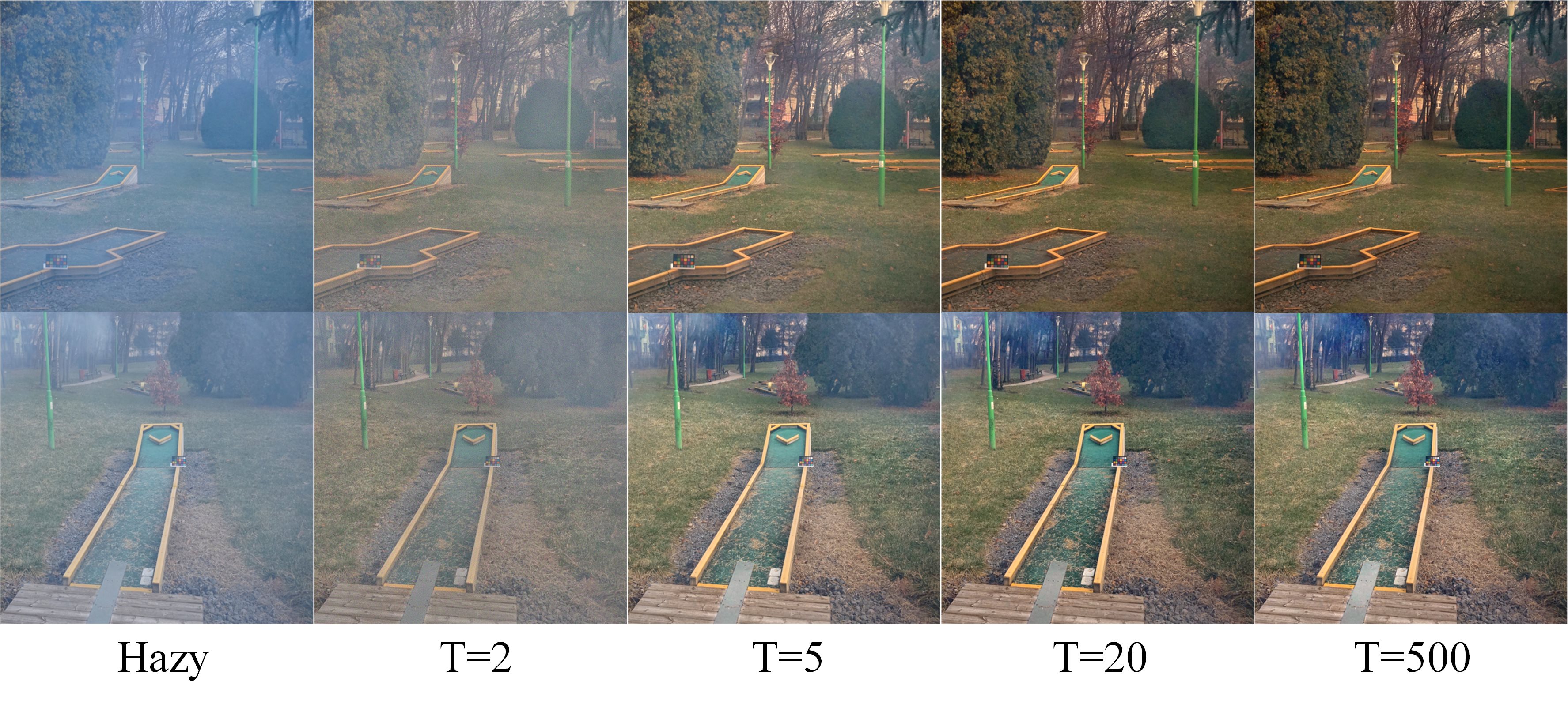}
	\caption{Influence of the sampling step $T$ on dehazing performance. The model achieves better trade-off at $T=20$.}
	\label{bstep}
\end{figure}

During evaluation, we adopt the DDIM framework to speed up sampling.
Fig.\ref{diff_dimm_aba} shows how PSNR and SSIM change with an increasing number of sampling step T, where we find a good trade-off between the performance and the sampling efficiency at $T=20$.
After that, the metrics are only slightly improved.
In this paper, all the experiment results are evaluated at $T=20$, but we note that even better results can be obtained with larger $T$.
In Fig.\ref{bstep}, we visually compare the sampling results and clearly see that the perceptual quality of the dehazed images is greatly improved by larger sampling steps, but starts to saturate when $T>20$.

\subsection{Effectiveness of the FCB during sampling}
\begin{figure}[!t]
	\centering
	\includegraphics [width=3.0in]{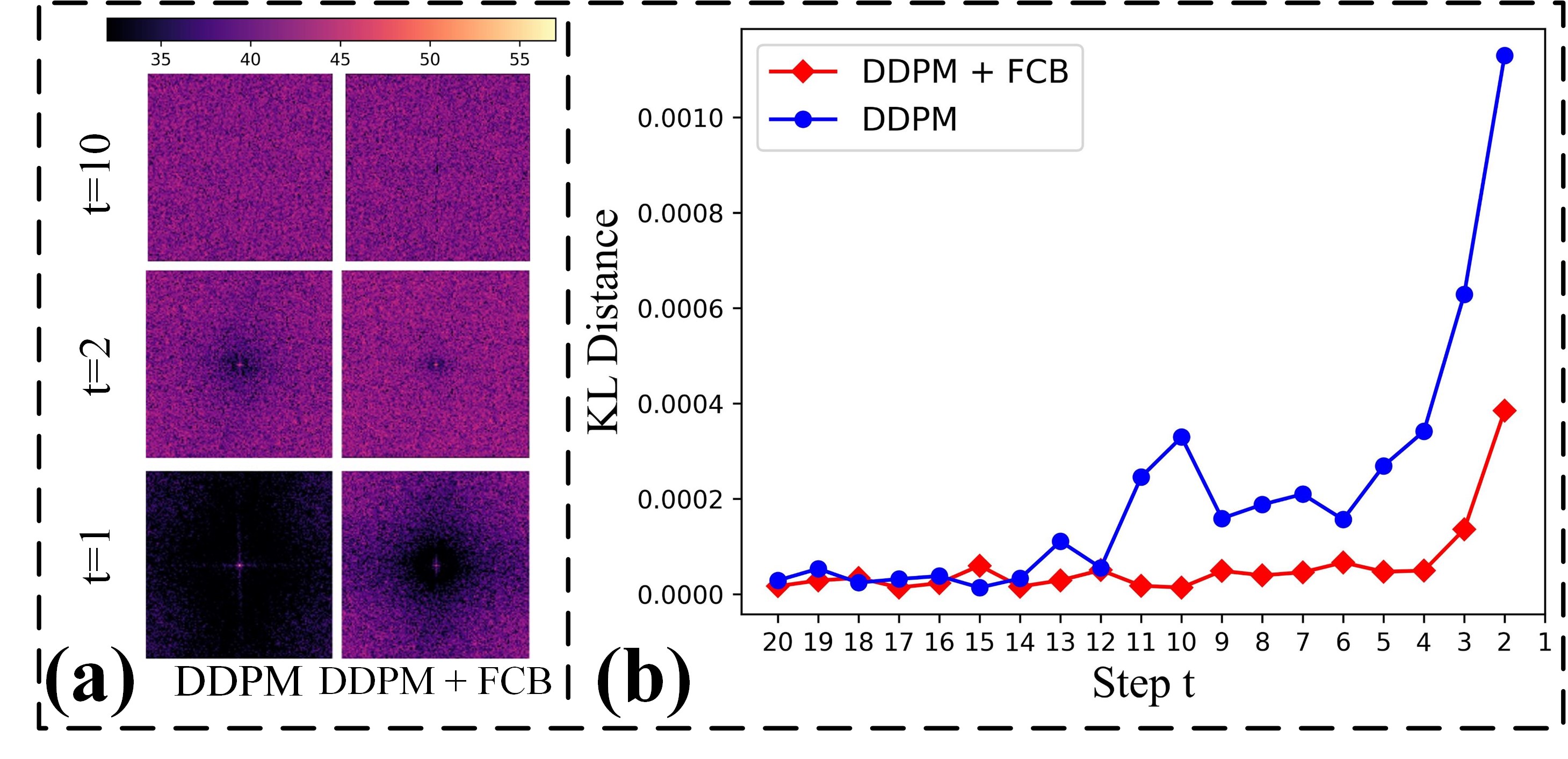}
	\caption{
Output power spectrum analysis during sampling.
(a) Comparison of the power spectra between the DDPM and DDPM+FCB.
(b) The KL distance between the PSD curves of the predicted $\epsilon$ and the Gaussian noise in (b).}
	\label{infer_prove}
\end{figure}

We also analyze the power spectrum of the outputs of diffusion models during sampling. Note that the sampling step is up to 20 due to the usage of DDIM.
Fig.\ref{infer_prove}a visually compares the power spectrum of the predicted $\epsilon$ between a plain DDPM and DDPM+FCB during sampling. 
As $t\rightarrow 1$,
the plain DDPM suffers from a lack of mid-to-high frequency modes, while our method significantly alleviates this problem.
In Fig.\ref{infer_prove}b, we further compute the KL distance between the PSD of the predicted $\epsilon$ and the Gaussian noise. The results indicate that FCB effectively pushes the output closer to an uniform distribution.

\begin{figure}[!t]
	\centering
	\includegraphics [width=2.8in]{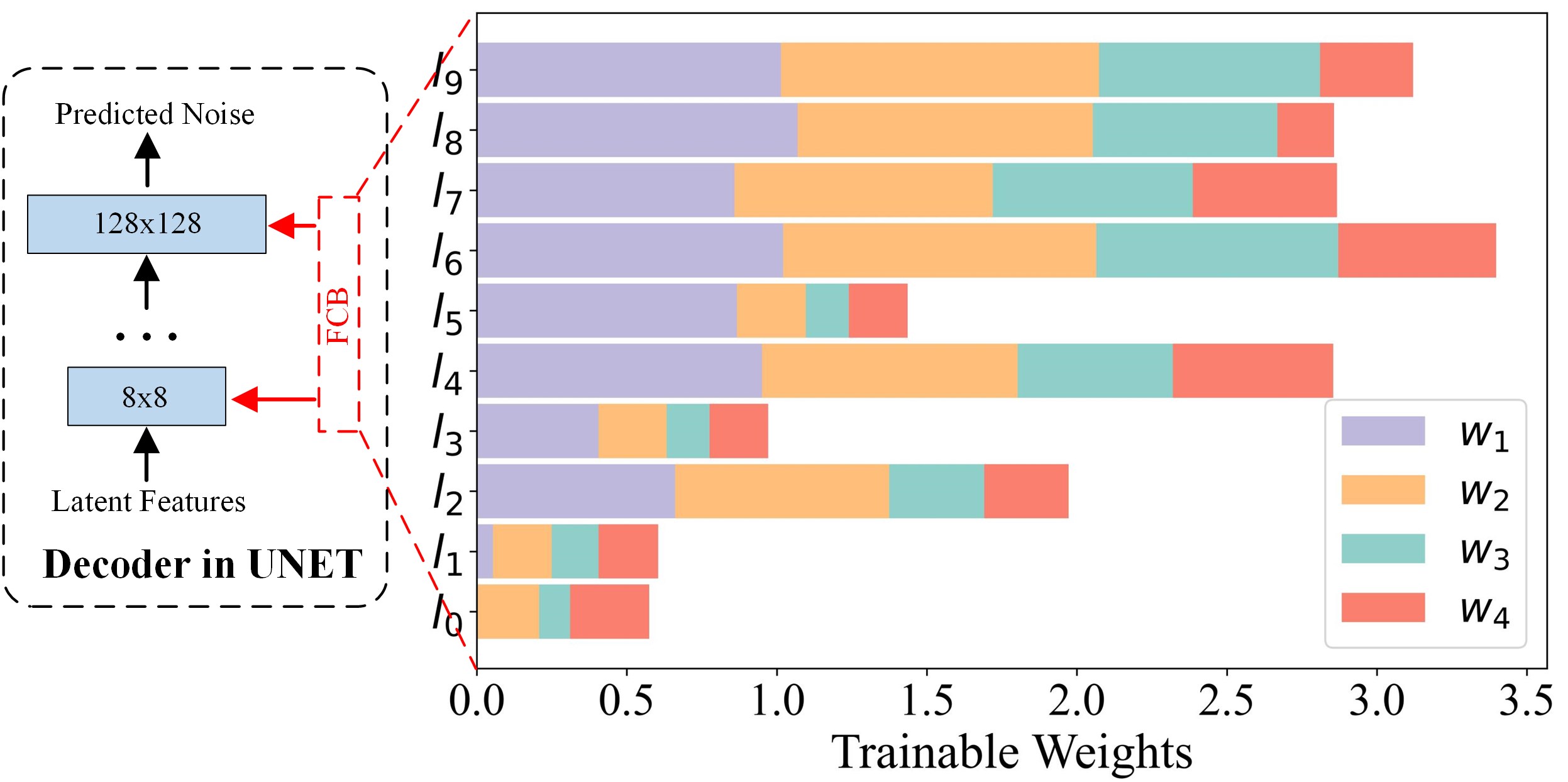}
	\caption{Visualization of the trainable weights $\mathbf{W}$ of different FCBs from $l_0$ to $l_9$. $l_9$ is the closest to the network output while $l_0$ is the most distant. $w_1$, $w_2$, $w_3$, $w_4$ are the weights of $s'$, $s'_3$, $s'_5$, $s'_3$ in each FCB, respectively. }
	\label{before_and_after_shortcut}
\end{figure}

\subsection{Additional ablation studies of FCB}
Firstly, We investigate the importance of the trainable weights $\mathbf{W}$ across all the 10 FCBs in a diffusion model. As shown in Fig.\ref{before_and_after_shortcut},
FCBs that are far away from the network output (\textit{e.g.,} $l_0$, $l_1$) tend to have small $\mathbf{W}$, which implies that the deeper FCBs benefit more from the mid-to-high frequency compensation. Within each FCB, we find that all the filters matter, even the identity mapping $w_1$, which further verifies the advantages of choosing $M>1$ in FCB design. 

\begin{table}[!t]
\scriptsize
\centering
\caption{Ablation experiments of FCB with different $\gamma_{\sigma}$. \textbf{Bold} and \uline{underline} indicate the best and the second-best, respectively.}
\resizebox{\linewidth}{!}{
\begin{tabular}{c|c|ccccc}
\hline
        Datasets        & $\gamma_{\sigma}$ & PSNR $\uparrow$ & SSIM $\uparrow$   & CIEDE $\downarrow$                              & MUSIQ $\uparrow$  & NIQE $\downarrow$          \\ \hline
\multirow{4}{*}{I-Haze} & 0.1  & 15.384  & 0.738     & 15.394                              & 62.571        & 4.639                   \\
                        & 0.5  & \textbf{16.379} & \textbf{0.766}      & 12.872                              & 65.491     & 4.429                 \\
                        & 1   & 15.983  & \uline{0.747}      & \multicolumn{1}{c}{\textbf{12.047}} & \multicolumn{1}{c}{{\uline{65.571}}} & \uline{4.408} \\
                        & 1.5  & \uline{16.342}  & 0.735     & {\uline{12.577}}                        & \textbf{65.662}     & \textbf{4.402}             \\ \hline
\multirow{4}{*}{O-Haze} & 0.1  & 15.874  & 0.648     & 16.771   & 61.013   & 2.827               \\
                        & 0.5  & 16.362 & 0.701      & 15.804          & 63.321     & \uline{2.555}                 \\
                        & 1   & \textbf{16.403}  & \textbf{0.712}        & \uline{15.805}  & \textbf{63.463}  & 2.556 \\
                        & 1.5  & \uline{16.377}  & \uline{0.708}     & \textbf{15.812}            & \uline{63.377}     & \textbf{2.553}             \\ \hline                        
\end{tabular}
}
\label{sigma_ratio}
\end{table}

Secondly, we explore how the cutoff frequency of the filters in FCB impacts the performance. To this end, $\sigma$ in Gaussian kernels are varied with respect to $\tilde{\sigma} = \gamma_{\sigma} \cdot \sigma$. By adjusting $\gamma_{\sigma}$, one can change the frequency response of the filters in FCB accordingly. For example, in our setting of $M=3$, enlarging $\gamma_{\sigma}$ consistently to all the Gaussian kernels will end up with a lower cutoff frequency and cover more low-frequency range.
As shown in Table \ref{sigma_ratio}, $\gamma_{\sigma}$ is adjusted from $\{ 0.1, 0.5, 1, 1.5 \}$ respectively. We can see that when the cutoff frequency is too high, \textit{e.g.,} $\gamma_{\sigma}=0.1$, the frequency range to be compensated is not large enough and hence results in a degraded performance. On the other hand, the performance can be maintained as long as the filter bank in FCB covers enough higher frequency band.

\subsection{HazeAug for other learning-based methods}
We also validate the effectiveness of HazeAug on other learning-based methods.
We select three representative methods (MSBDN, GCA and DehazeFormer) that perform very well on synthetic data, and apply HazeAug to them (see Table \ref{GCA-HazeAug}, Table \ref{MSBDN-HazeAug}, and Table \ref{DF-HazeAug} respectively).
We notice that even applied with HazeAug, the performances of MSBDN, GCA and DehazeFormer are still inferior to our method (see Table \textcolor{red}{1}) on all the metrics except MUSIQ (DehazeFormer is better at MUSIQ on O-Haze and Dense-Haze). On the reference-based metrics, in particular, they are largely outperformed by our method, which once again validates the effectiveness of the proposed frequency compensation module.

In Table \ref{GCA-HazeAug}, Table \ref{MSBDN-HazeAug} and Table \ref{DF-HazeAug}, we also observe that  HazeAug is very effective in improving MUSIQ compared to traditional augmentation, but having no consistent boosts on the other metrics. For example, HazeAug shows little impact on SSIM, while in the case of CIEDE, the performances even decrease. In contrast, our method (see Table \textcolor{red}{3}) results in consistent improvements across these metrics, especially with more significant performance \textit{gains} (\textit{e.g.,} +0.076 vs. +0.005 for SSIM and +3.221 vs. +0.860 for MUSIQ once compared to MSBDN on I-Haze). We attribute it to that the diffusion models possess more powerful data modeling ability, which can be unleashed by sufficiently augmented data. As HazeAug effectively increases haze diversity and imlicitly encourages generative modeling, diffusion models demonstrate excellent scalability to richer dataset, thus generalizing better to real images than the end-to-end learning methods. 


\begin{table}[!t]
\scriptsize
\centering
\caption{Performance of GCANet with HazeAug. \textbf{Blod} denotes the results that are improved by HazeAug. \textcolor{red}{Red} denotes the performance that has \textcolor{red}{decreased} compared to the traditional augmentation, and \textcolor{blue}{Blue} denotes the performance that has \textcolor{blue}{increased}.}
\resizebox{\linewidth}{!}{
\begin{tabular}{c|ccccc}
\hline
     Datasets     & PSNR $\uparrow$         & SSIM $\uparrow$         & CIEDE $\downarrow$ & MUSIQ $\uparrow$         & NIQE $\downarrow$ \\ \hline
\multirow{1}{*}{I-Haze} &  15.852 \textcolor{red}{(-1.127)}           & 0.711 \textcolor{red}{(-0.035)}         & 24.198 \textcolor{red}{(+0.306)}         & \textbf{64.929} \textcolor{blue}{(+2.811)} & \textbf{4.533} \textcolor{blue}{-0.538} \\
\hline
\multirow{1}{*}{O-Haze}      & 13.529 \textcolor{red}{(-1.026)}          & \textbf{0.766} \textcolor{blue}{(+0.104)} & 25.129 \textcolor{red}{(+0.060)}   & \textbf{65.547} \textcolor{blue}{(+2.159)} & \textbf{3.550} \textcolor{blue}{(-0.267)} \\
 \hline
\multirow{1}{*}{Dense-Haze} &\textbf{11.948}  \textcolor{blue}{(+1.896)} & \textbf{0.464}  \textcolor{blue}{(+0.077)} & \textbf{23.702} \textcolor{blue}{(-3.817)} & \textbf{43.555}  \textcolor{blue}{(+1.935)}  & \textbf{4.953} \textcolor{blue}{(-1.750)} \\
\hline
\multirow{1}{*}{Nh-Haze}   & \textbf{12.160} \textcolor{blue}{(+0.528)} & \textbf{0.530} \textcolor{blue}{(+0.037)}  & 22.047 \textcolor{red}{(-0.391)}          & \textbf{63.367} \textcolor{blue}{(+1.168)} & \textbf{3.320} \textcolor{blue}{(-0.089)} \\ \hline
\end{tabular}
}
\label{GCA-HazeAug}
\end{table}

\begin{table}[!t]
\scriptsize
\centering
\caption{Performance of MSBDN with HazeAug. \textbf{Blod} denotes the results that are improved by HazeAug. \textcolor{red}{Red} denotes the performance that has \textcolor{red}{decreased} compared to the traditional augmentation, and \textcolor{blue}{Blue} denotes the performance that has \textcolor{blue}{increased}.}
\resizebox{\linewidth}{!}{
\begin{tabular}{c|ccccc}
\hline
     Datasets      & PSNR $\uparrow$         & SSIM $\uparrow$         & CIEDE $\downarrow$ & MUSIQ $\uparrow$         & NIQE $\downarrow$ \\ \hline
\multirow{1}{*}{I-Haze}   & 16.246 \textcolor{red}{(-0.742)}          & \textbf{0.753} \textcolor{blue}{(+0.005)}         & 24.673 \textcolor{red}{(+0.356)}   & \textbf{64.200} \textcolor{blue}{(+0.860)} & 6.944 \textcolor{blue}{(+1.516)} \\ \hline
\multirow{1}{*}{O-Haze}      
     & 16.537  \textcolor{red}{(-0.025)}        & \textbf{0.684} \textcolor{blue}{(+0.033)} & 25.748 \textcolor{red}{(+2.464)} & \textbf{63.900} \textcolor{blue}{(+0.743)} & 7.108 \textcolor{red}{(+2.643)} \\ \hline
\multirow{1}{*}{Dense-Haze} 
& \textbf{12.436} \textcolor{blue}{(+1.292)} & \textbf{0.442} \textcolor{blue}{(+0.050)} & 28.716 \textcolor{red}{(+3.781)} & \textbf{42.058} \textcolor{blue}{(+0.413)} & 7.401 \textcolor{red}{(+0.558)} \\ 
\hline
\multirow{1}{*}{Nh-Haze}   & 12.008 \textcolor{red}{(-0.492)}         & \textbf{0.567} \textcolor{red}{(-0.09)} & 26.668 \textcolor{red}{(+5.337)} & \textbf{63.023} \textcolor{blue}{(+1.058)} & 5.523 \textcolor{red}{(+1.738)} \\
\hline
\end{tabular}
}
\label{MSBDN-HazeAug}
\end{table}

\begin{table}[!t]
\scriptsize
\centering
\caption{Performance of DehazeFormer with HazeAug. \textbf{Blod} denotes the results that are improved by HazeAug. \textcolor{red}{Red} denotes the performance that has \textcolor{red}{decreased} compared to the traditional augmentation, and \textcolor{blue}{Blue} denotes the performance that has \textcolor{blue}{increased}.}
\resizebox{\linewidth}{!}{
\begin{tabular}{c|ccccc}
\hline
     Datasets     & PSNR $\uparrow$         & SSIM $\uparrow$         & CIEDE $\downarrow$ & MUSIQ $\uparrow$         & NIQE $\downarrow$ \\ \hline
\multirow{1}{*}{I-Haze} & \textbf{16.610}  \textcolor{blue}{(+0.060)} & 	0.733  \textcolor{red}{(-0.021)}	&24.068 \textcolor{red}{(+11.501)} &	64.222 \textcolor{red}{(-0.739)}& 	7.358  \textcolor{red}{(+2.980)}
 \\
\hline
\multirow{1}{*}{O-Haze}     & 15.042 \textcolor{red}{(-1.298)} &	0.667 \textcolor{red}{(-0.033)}  &	24.104 \textcolor{red}{(+8.207)} 	 &\textbf{65.874} \textcolor{blue}{(+1.966)}	 & 6.973 \textcolor{red}{(+3.919)}  \\
 \hline
\multirow{1}{*}{Dense-Haze} & \textbf{11.557} \textcolor{blue}{(+0.707)} &	\textbf{0.416} \textcolor{blue}{(+0.035)} & 	\textbf{24.803} \textcolor{blue}{(-0.721)} &	\textbf{44.984} \textcolor{blue}{(+2.901)} &	7.959 \textcolor{blue}{(-1.568)}  \\\hline
\multirow{1}{*}{Nh-Haze}    &\textbf{12.543}  \textcolor{blue}{(+0.293)}  &	\textbf{0.542}  \textcolor{blue}{(+0.044)} &	\textbf{21.010}  \textcolor{blue}{(-0.614)}  &	\textbf{63.760}  \textcolor{blue}{(+0.761)}  &	5.490  \textcolor{red}{(+2.256)}
 \\ \hline
\end{tabular}
}
\label{DF-HazeAug}
\end{table}

\section{Additional Comparison Results}

\subsection{More visualization results}
\begin{figure*}[!t]
	\centering
	\includegraphics [width=6.0in]{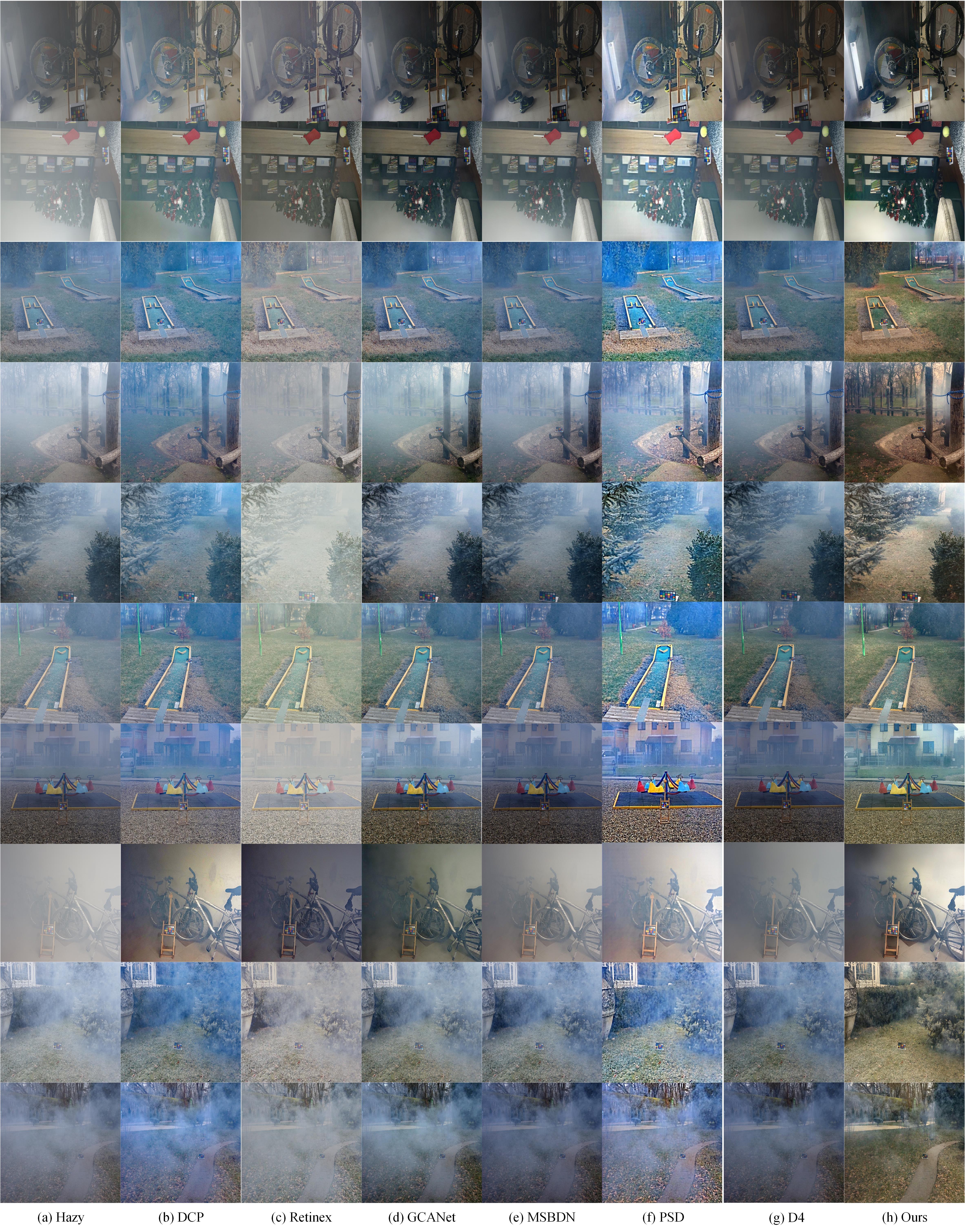}
	\caption{
	Qualitative comparison of different methods on I-Haze, O-Haze, Dense-Haze, and Nh-Haze.
	}
	\label{appdix2}
\end{figure*}

\begin{figure*}[!t]
	\centering
	\includegraphics [width=6.2in]{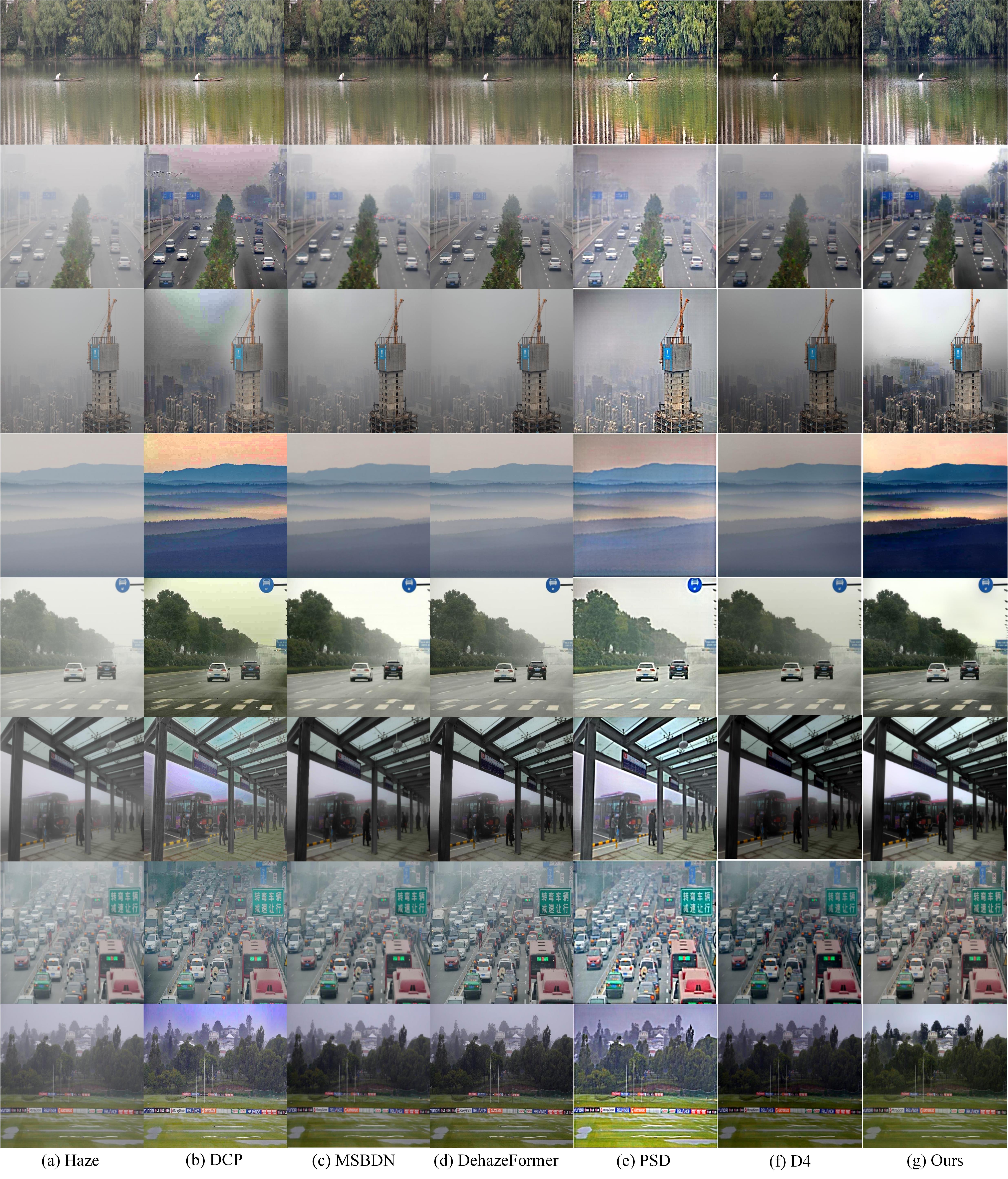}
	\caption{
	Qualitative comparison of different methods on the RTTS dataset.
	}
	\label{appdix1}
\end{figure*}

We report more visualization results of our proposed method.
We first compare the visual outputs on I-Haze, O-Haze, Dense-Haze, and Nh-Haze in Fig.\ref{appdix2}, and then on real-world RTTS database in Fig.\ref{appdix1}. 
The dehazed images by our method are more natural and have less haze residue, which verifies the effectiveness of the proposed dehazing framework on real-world hazy images.

